\documentclass[10pt,twocolumn,letterpaper]{article}

\usepackage[pagenumbers]{cvpr} 

\pdfoutput=1

\usepackage{bm}
\usepackage{multirow}

\definecolor{cvprblue}{rgb}{0.21,0.49,0.74}
\usepackage[pagebackref,breaklinks,colorlinks,allcolors=cvprblue]{hyperref}

\usepackage[all=normal, mathspacing=normal,mathdisplays=normal, floats=tight, paragraphs=tight, lists=normal]{savetrees}
\usepackage{comment}

\def\paperID{1965} 
\def\confName{CVPR}
\def\confYear{2025}

\newcommand{\method}{PUNC}

\def\vtheta{{\bm{\theta}}}

\def\vc{{\bm{c}}}

\def\vx{{\bm{x}}}

\def\vz{{\bm{z}}}

\title{Towards Understanding and Quantifying Uncertainty\\for Text-to-Image Generation}
\author{\textbf{Gianni Franchi},\textsuperscript{\rm 1}  \textbf{Dat Nguyen Trong},\textsuperscript{\rm 1}  \textbf{Nacim Belkhir},\textsuperscript{\rm 2} \\
\textbf{Guoxuan Xia},\textsuperscript{\rm 3} 
\textbf{Andrea Pilzer}\textsuperscript{\rm 4}  \\
U2IS, ENSTA Paris, Institut Polytechnique de Paris, \textsuperscript{\rm 1}Mirai,\textsuperscript{\rm 2}  \\
Imperial College London,\textsuperscript{\rm 3}  NVIDIA\textsuperscript{\rm 4}
}

\begin{document}

\twocolumn[{
\renewcommand\twocolumn[1][]{#1}%
\maketitle
\begin{center}
    \centering
    \captionsetup{type=figure}
    \captionsetup[subfigure]{labelformat=empty}
    \addtocounter{figure}{-1}
    \begin{subfigure}[b]{0.99\textwidth}
         \centering
         \includegraphics[width=0.99\textwidth]{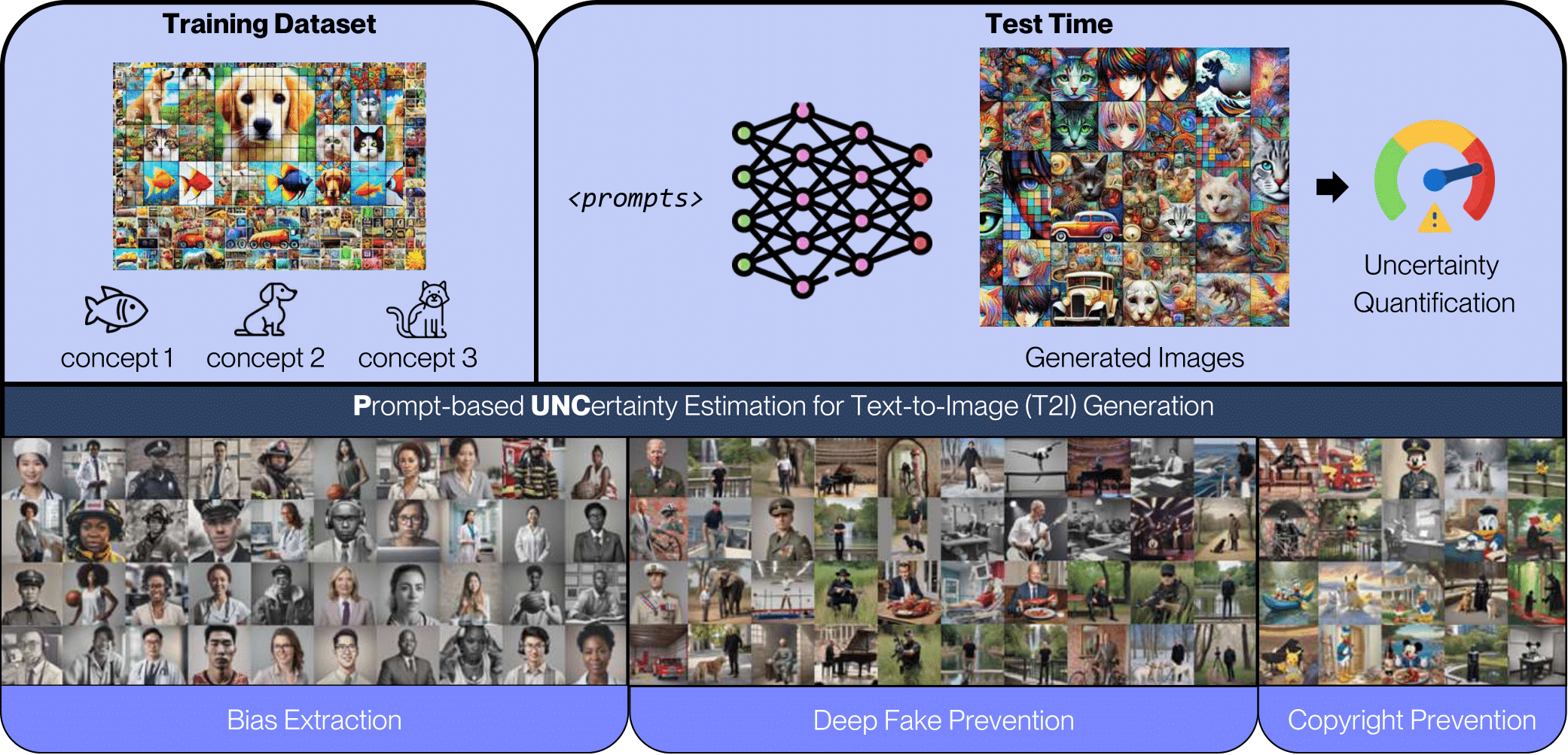}
    \end{subfigure}
          \hfill
    \captionof{figure}{\textbf{Examples of Applications for Uncertainty Quantification in Text-to-Image Generation.} Text-to-image generation models may exhibit uncertainty, and that need to be quantified since it can provide insights into the model’s training dataset, aiding in deepfake prevention, detecting model biases, and protecting copyrighted content from unauthorized generation.}
    \label{fig:qualitative_robustness}
\end{center}}]

\begin{abstract}

Uncertainty quantification in text-to-image (T2I) generative models is crucial for understanding model behavior and improving output reliability. In this paper, we are the first to quantify and evaluate the uncertainty of T2I models \textit{with respect to the prompt}. Alongside adapting existing approaches designed to measure uncertainty in the image space, we also introduce \textbf{\underline{P}rompt-based \underline{UNC}ertainty Estimation for T2I models (\method)}, a novel method leveraging Large Vision-Language Models (LVLMs) to better address uncertainties arising from the \textit{semantics} of the prompt and generated images. \method~ utilizes a LVLM to caption a generated image, and then compares the caption with the original prompt in the more semantically meaningful \textit{text} space. PUNC also enables the disentanglement of both aleatoric and epistemic uncertainties via precision and recall, which image-space approaches are unable to do. 
Extensive experiments demonstrate that \method~ outperforms state-of-the-art uncertainty estimation techniques across various settings. Uncertainty quantification in text-to-image generation models can be used on various applications including bias detection, copyright protection, and OOD detection. We also introduce a comprehensive dataset of text prompts and generation pairs to foster further research in uncertainty quantification for generative models. Our findings illustrate that \method~ not only achieves competitive performance but also enables novel applications in evaluating and improving the trustworthiness of text-to-image models.
\end{abstract}    
\section{Introduction}
\label{sec:intro}

Generative models have revolutionized artificial intelligence, making significant strides in fields such as robotics \cite{ahn2022can}, healthcare~\cite{jumper2021highly}, and physics~\cite{lam2023learning}. These models have elevated the performance of AI systems by enabling the generation of complex data representations. However, while generative models offer immense potential, they do not address all challenges, especially those related to robustness and uncertainty in Deep Neural Networks(DNNs)—issues that have become even more pronounced in generative settings.

In this paper, we focus on the \textbf{uncertainty quantification} in deep generative models, in particular, we are the first to investigate modern \textbf{text-to-image (T2I) models} where the input is a natural language prompt \cite{podellsdxl,chen2024pixartalpha,sd3,rombach2022high,dalle}. These models have surged in popularity in recent years due to their capacity to generate high-quality images conveniently in a user-guided fashion, finding usage in a wide array of real-world applications. We are the first to investigate and propose a dedicated approach for quantifying uncertainty in text-to-image generation, aiming to fill a critical gap in generative AI research.
%
The task of quantifying uncertainty in these models is not only complex but also crucial. Understanding the uncertainty in generative models can be invaluable for applications such as Out-of-Distribution (OOD) detection, trustworthy AI, and decision-making in high-stakes domains. To the best of our knowledge, no existing work has comprehensively addressed the challenge of uncertainty quantification in text-to-image generative models.

Existing research explores uncertainty estimation and OOD detection in the \textit{image/output} space \cite{graham2023denoising,liu2023unsupervised,earthobs,ensdiffdecu}. For example, given a test image, can we determine whether or not it is OOD? On the other hand, the \textit{uncertainty of a generation with respect to conditioning} (such as text) is underexplored. We argue that this is an important area to be addressed, as it represents a widespread real-world application of deep generative models. In this case, it remains an open question as to \textit{how should we define uncertainty}, and given a definition, \textit{how should we quantify it}? As illustrated in \cref{fig:qualitative_robustness}, focusing on uncertainty quantification in text-to-image generation could enable novel applications, allowing us to extract and interpret the knowledge embedded within models using language.

To answer these questions, we first argue that uncertainties should be rooted in the \textit{semantics} of the image and the prompt. We then adapt and evaluate a number of existing \textit{image-space} approaches, as well as proposing a novel method for uncertainty quantification in T2I generative models, which we refer to as ``\underline{P}rompt-based \underline{UNC}ertainty  Estimation for T2I Generation'' (\method). Our approach leverages the growing capabilities of Large Vision-Language Models (LVLMs) to extract the semantics from generated images. LVLMs, which have been trained on vast amounts of text and visual data, can serve as powerful tools for interpreting the underlying meaning of text/image inputs and assessing how variations in phrasing or semantics impact the uncertainty of the generated outputs. Our hypothesis is that as LVLMs become increasingly adept at understanding complex relationships between text and images, they will allow us to better analyze and quantify uncertainty. Through this paper, we will demonstrate Prompt-based Uncertainty Estimation for T2I (\method) effectiveness in quantifying uncertainties in generative models. Showcasing its ability to decompose and interpret different types of uncertainties arising from the multimodal nature of the input, offering a new perspective on uncertainty quantification in generative AI.



To summarize, our contributions are as follows:
\textbf{(1)} We introduce a new task aimed at quantifying uncertainty (with respect to the prompt) in text-to-image generation models, and we share a dataset of prompts to support future research on uncertainty in these models.  
\textbf{(2)} We both adapt existing image-space approaches to our tasks, as well as propose a novel straightforward method, \method, designed to better quantify semantic uncertainty in text-to-image generation.
\textbf{(3)} Extensive experiments demonstrate that \method, despite its simplicity and computational efficiency, achieves competitive performance.
\textbf{(4)} We highlight the practical utility of uncertainty quantification in text-to-image generation by showcasing additional applications where this approach could be beneficial.

\section{Related Work}
\label{sec:related}

Deep learning models typically encounter two primary types of uncertainty: aleatoric and epistemic uncertainty \cite{H_llermeier_2021}. Over the years, a variety of methods have been developed to quantify these uncertainties. Ensemble methods \cite{lakshminarayanan2017simple,wen2019batchensemble,durasov2021masksembles,laurent2023packed,window-casc}, for instance, have often achieved state-of-the-art performance by leveraging multiple models to assess prediction confidence. However, these methods tend to be computationally expensive due to the need to train and maintain several models. Another approach involves Bayesian methods \cite{Galthesis}, especially Bayesian Neural Networks (BNNs) \cite{neal2011mcmc,izmailov2021bayesian,welling2011bayesian,franchi2020tradi}, which offer a theoretically grounded way to quantify uncertainty. Additionally, there are techniques aimed at directly predicting uncertainty within DNNs \cite{kendall2017uncertainties,corbiere2019addressing,LDU}. Despite their effectiveness, most of these methods are challenging to apply to text-to-image generation, as they rely on ensembling predictions or accessing model uncertainty in ways that are difficult to implement for generative models without explicit access to intermediate predictions.

Generative models, as a subset of deep learning, also exhibit uncertainty. In the past, traditional generative models were employed for uncertainty quantification and used to help estimate the uncertainty within standard DNNs \cite{lis2019detecting,vojir2021road,xia2020synthesize,chan2024hyper}. With the recent popularity of diffusion models, research has begun exploring whether image-based diffusion models can detect out-of-distribution (OOD) samples \cite{graham2023denoising,liu2023unsupervised,earthobs} or quantify their own uncertainty \cite{chan2024hyper,du2023diffusion,ensdiffdecu}. Whilst these approaches align with our goals, they are limited to image-space tasks, where the input is an image rather than a textual prompt. This input difference fundamentally alters the nature of uncertainty in the model. Our approach, in contrast, specifically addresses text-to-image generation, a setting where prompt ambiguity and multimodal complexity introduce new dimensions of uncertainty. We demonstrate that quantifying uncertainty in this context can lead to valuable insights and applications across various downstream tasks.

%
%

\section{Uncertainty in Image Generation Models}
\label{sec:uncertainty}

\begin{figure}[t]
	\centering
    \includegraphics[width=0.9\columnwidth]{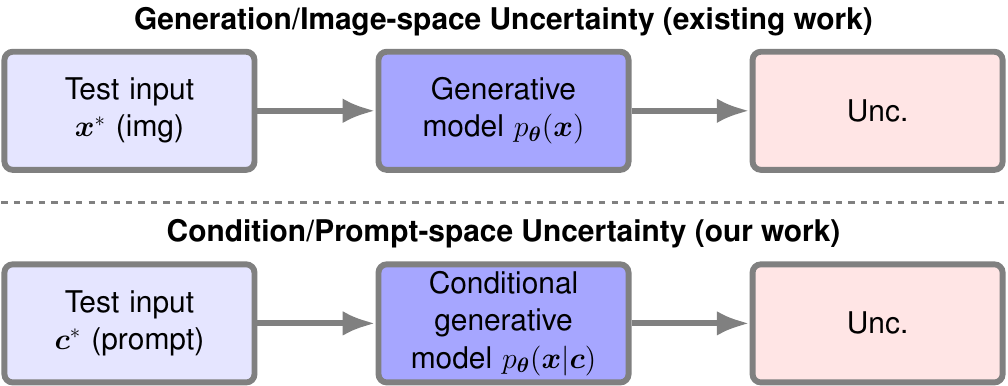}
    \caption{Diagram illustrating generation/image-space uncertainty which is considered in previous work and condition/prompt-space uncertainty which is investigated in our work.}
    \label{fig:uncertainty_diagram}
\end{figure}

\begin{figure}[t]
\centering
\resizebox{0.95\columnwidth}{!}
{
\begin{tabular}{ccc}
Normal & Corruption & OOD  \\
\includegraphics[width=0.3\columnwidth]{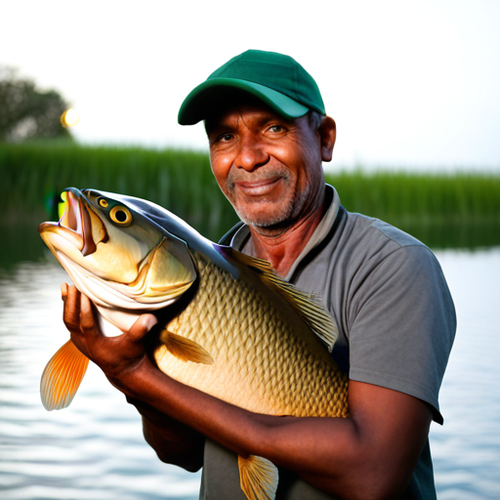}&
\includegraphics[width=0.3\columnwidth]{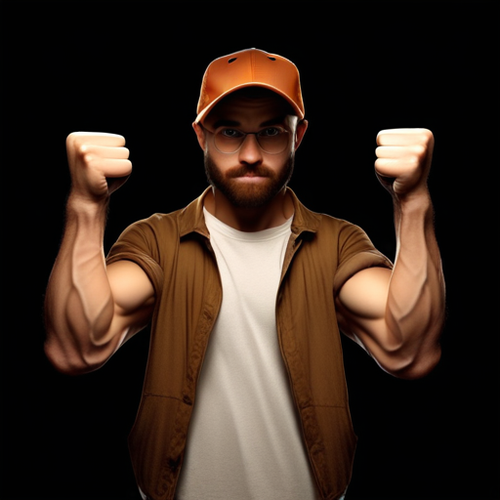}&
\includegraphics[width=0.3\columnwidth]{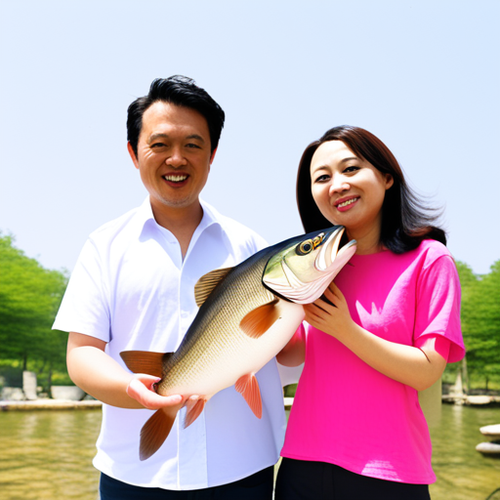}
\end{tabular}
}
\caption{Three generations from PixArt-$\Sigma$ \cite{chen2024pixart} illustrating uncertainty with regards to prompt \textit{semantics}. For the Normal image, we used an ImageNet-inspired prompt; for the corrupted image, additional corruption was applied to the prompt (\eg \texttt{fish} was perturbed to \texttt{fis}) increasing the \textit{aleatoric} uncertainty; and for the out-of-distribution (OOD) case, the model was prompted to generate an image of the Prime Minister of Japan \texttt{Kishida Fumio} under similar conditions, where \textit{epistemic} uncertainty was injected in the form of a semantic concept the model was not familiar with.}
\vspace{-1em}
\label{fig:image_illustration_uncertainty}
\end{figure}

Consider a deep generative model $p_\vtheta(\vx)$ with parameters $\vtheta$ that models the data distribution $p_\text{data}(\vx)$ of $\vx \in \mathbb{R}^D$. We may be interested in measuring the uncertainty of our model $p_\vtheta$ with respect to some test datum $\vx^*$. For example, consider the case where we want to determine whether or not $\vx^*$ was drawn from $p_\text{data}(\vx)$, \ie out-of-distribution (OOD) detection. This can be achieved by quantifying the uncertainty of $p_\vtheta$ that arises from a\textit{ lack of knowledge} of $\vx^*$, \ie \textit{epistemic uncertainty} \cite{H_llermeier_2021}.\footnote{We note that although there is a discussion over how to define and quantify epistemic uncertainty \cite{xia2022usefulnessdeepensemblediversity,pearce2021understandingsoftmaxconfidenceuncertainty,priornets,Galthesis,H_llermeier_2021}, in this work, we stick to an intuitive albeit informal view based on a deep model's \textit{knowledge}.} A range of existing work that leverages diffusion-based generative models for OOD image detection \cite{liu2023unsupervised,graham2023denoising} fall into the above paradigm. From hereon we will refer to this as generation/image-space uncertainty. This is illustrated in \cref{fig:uncertainty_diagram} (top).

In this work, however, we are primarily interested in a different scenario: text-to-image generation. Consider instead a \textit{conditional} model $p_\vtheta(\vx|\vc)$ that models the conditional data distribution $p_\text{data}(\vx|\vc)$. This describes real-world text-to-image generation, where image $\vx$ is drawn from $p_\vtheta(\vx|\vc^*)$ conditioned on some test text prompt $\vc^*$. In this case, we wish to quantify the uncertainty of $p_\vtheta$ not with respect to some test $\vx^*$ (\eg image) but instead with respect to a test \textit{condition} $\vc^*$. We will hereon refer to this as condition/prompt-space uncertainty. This is illustrated in \cref{fig:uncertainty_diagram} (bottom).

We argue that in this case, uncertainties should be concerned with \textit{semantics}, as intuitively, these are what inherently link the \textit{user's intentions}, represented by the text prompt, and the generated output image. This is similar to recent discourse on uncertainty estimation for large language models \cite{semantic_unc}. If we consider the semantic concepts in the prompt and generated image, then we can define useful concepts of uncertainty (illustrated in \cref{fig:image_illustration_uncertainty}):
\begin{itemize}
    \item \textbf{Aleatoric uncertainty} is irreducible uncertainty in the data distribution $p_\text{data}(\vx|\vc)$. From the perspective of semantics, high aleatoric uncertainty would be where a variation of generated concepts may arise from a single prompt. For example, a spelling mistake where \texttt{fish} is mistyped as \texttt{fis} in the prompt may result in either a ``fish'' or a ``fist'' appearing in the image.
    \item \textbf{Epistemic uncertainty} should correspond to a model's \textit{lack of knowledge} of semantic concepts found in a prompt. For example, a model trained on ImageNet will not know what the prime minister of Japan \texttt{Kishida Fumio} looks like, and thus will have high epistemic uncertainty for this semantic concept.
\end{itemize}
We note that broadly speaking uncertainty estimation for deep conditional models is well explored \cite{hendrycks2017a,priornets,xia2022usefulnessdeepensemblediversity,LDU,scod,window-casc,lakshminarayanan2017simple,Galthesis,H_llermeier_2021,oodsurvey,ls_sc,deep_sc,laurent2023packed}. However, research in this area is generally concerned with problem settings where 1) the models \textit{explicitly} evaluate $p_\vtheta(\vx|\vc)$ and 2) downstream applications of uncertainty relate directly to the numerical values of $\vx$ and $p_\vtheta(\vx|\vc)$. For example, a cross-entropy-trained image classifier explicitly predicts the probability mass $P_\vtheta(\vx|\vc^*)$ of label $\vx$ given test image $\vc^*$.\footnote{We stick with this non-standard notation for the sake of consistency.} $P_\vtheta(\vx|\vc^*)$ directly relates to the likelihood of classification error -- which is an event that we may want to detect using an uncertainty measure, \ie downstream misclassification detection \cite{hendrycks2017a,deep_sc}. Thus, we are able to easily extract useful measures of uncertainty from the model $P_\vtheta$, \eg the maximum of the softmax. 

On the other hand, text-to-image generation models are not as clear-cut. They may not explicitly evaluate $p_\vtheta(\vx|\vc^*)$ (\eg diffusion-based models \cite{ho2020denoising,song2021scorebased}), and even when they do (\eg vector-quantised autoregressive models \cite{VAR,taming}) it is hard to directly relate (latent) pixel values directly to our above definitions of uncertainty. For example, simple variations in contrast and brightness that may leave semantics completely unchanged could result in a very high variance in $p_\vtheta(\vx|\vc^*)$. To give a simple example, inverting the black and white stripes on a zebra would lead to a large Euclidean distance in the image space whilst preserving the semantics of the image. In the next section, we present and discuss a range of potential methods to quantify the uncertainty of text-to-image models with respect to a given prompt.

\section{Techniques to Quantify Uncertainty}
\begin{figure*}[t]
\centering
\includegraphics[width=1.00\linewidth]{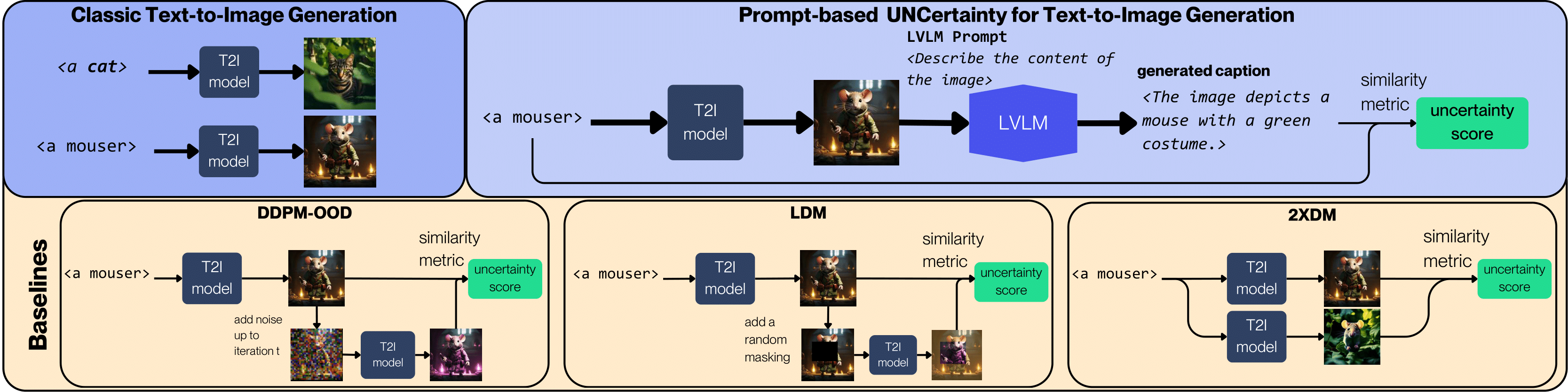}
\caption{\textbf{Illustration showing the different baselines and \method.} \method~ leverages a LVLM to describe generated images and assess similarity with the original prompt, providing a refined uncertainty score. In contrast, baseline methods employ traditional techniques such as noise injection, ensembling, or masking to quantify uncertainty, followed by image-based similarity scoring.
}
\vspace{-1em}
\label{fig:image_method}
\end{figure*}

In order to quantify the uncertainty of text-to-image generation models we will both adapt a number of existing approaches as well as propose a novel approach which we refer to as \underline{P}rompt-based \underline{UNC}ertainty estimation (PUNC). All methods are illustrated in \cref{fig:image_method}.
\subsection{Adapting Existing Approaches}\label{sec:baseline_methods}



There are a number of existing methods for quantifying the \textit{image-space} uncertainty of diffusion models. As diffusion-based models represent a majority of current text-to-image foundation models \cite{sd3,rombach2022high,chen2024pixartalpha,podellsdxl}, we choose to adapt these in order to quantify \textit{prompt-space} uncertainty (\cref{fig:image_method}). We first present a brief primer on diffusion models to the reader.


\subsubsection{Diffusion Models}
\renewcommand{\b}{\boldsymbol}
Diffusion models \cite{ho2020denoising} are a powerful class of generative model that generate data by starting from pure noise and then performing iterative denoising.
We define a \textit{forward process} over time \( t \in [0, 1] \) for the random variable \( \b x_t \in \mathbb{R}^D \), conditioned on a text prompt \( \b{c} \),
\begin{equation}\label{eq:forward}
    p(\b x_t | \b x_0, \b c) = \mathcal{N}(\b x_t; a_t \b x_0, \sigma_t^2 \b I)
\end{equation}
where \( \b x_0 \) is sampled from an unknown data distribution \( p(\b x_0) \). The functions \( a_t \) and \( \sigma_t \) define the \textit{noise schedule}, which controls the level of noise added over time. A stochastic differential equation (SDE) that shares the same conditional distributions as \cref{eq:forward} is given by \cite{kingma2021on}:
\begin{equation}\label{eq:sde}
    d\b x_t = f_t \b x_t dt + g_t d\b w_t, \quad \b x_0 \sim p(\b x_0),
\end{equation}
where \( \b w_t \in \mathbb{R}^D \) represents a Wiener process. The drift and diffusion coefficients are defined as
\begin{equation}
    f_t = \frac{d a_t}{dt} \frac{1}{a_t}, \quad g_t^2 = \frac{d \sigma_t^2}{dt} - 2 f_t \sigma_t^2.
\end{equation}
Given a neural network \( \b s_\theta(\b x_t, t, \b c) \) that approximates the score \( \nabla_{\b x_t} \log p(\b x_t | \b c) \), we can approximately generate samples from \( p(\b x_0 | \b c) \), starting from pure Gaussian noise, by either solving the \textit{reverse-time SDE} \cite{song2021scorebased},
\begin{equation}\label{eq:reverse_sde}
    d\b x_t = \left[ f_t \b x_t - g_t^2 \nabla_{\b x_t} \log p(\b x_t | \b c) \right] dt + g_t d\bar{\b w}_t,
\end{equation}
where \( \bar{\b w}_t \) is a reverse-time Wiener process, or by solving the \textit{probability flow ordinary differential equation (ODE)}:
\begin{equation}\label{eq:ode}
    \frac{d\b x_t}{dt} = f_t \b x_t - \frac{1}{2} g_t^2 \nabla_{\b x_t} \log p(\b x_t | \b c), \quad \b x_1 \sim p(\b x_1).
\end{equation}
 Various SDE and ODE solvers/samplers can be leveraged to generate high-quality samples via iterative denoising with few integration steps \cite{dpm_solver,ddim,deis,deis_sn,xu2023restart,song2021scorebased,elucidate}. 




Latent diffusion models (LDMs) \cite{liu2023unsupervised} perform this process within a compressed latent space rather than directly on pixel space. By first encoding images into a more compact representation, LDMs reduce computational demands, enabling high-quality high-resolution generations. We will not consider the latent representation to keep the notation simple, even though experiments will use LDMs.


\subsubsection{Time-step based approaches}

\paragraph{DDPM-OOD \cite{graham2023denoising}.} This approach, developed for image-space OOD detection, involves noising a test sample $\b x^*_0$ to different points in the forward process to get $\{\b{x}^*_{t_1}, \b{x}^*_{t_2},\dots\}$ and denoising them, generating multiple reconstructions $\{\hat{\b{x}}_{0_1}, \hat{\b{x}}_{0_2},\dots\}$. Similarity is then measured between each reconstruction and the original image. The idea is that test samples $\b x^*_0$ for which the model has higher epistemic uncertainty will lead to reconstructions that differ from the test sample. However, in our case, we do not have a test image $\b x^*_0$. In order to adapt DDPM-OOD we propose measuring similarity between reconstructions at different $t$ and a baseline generation, \(S(\hat{\b{x}}_{0_i}, \hat{\b{x}}_{0})\), where \(\hat{\b{x}}_{0}\) represents an image directly generated using the original test prompt $\b c^*$ alone. Both aleatoric uncertainty (\eg ambiguous prompt) and epistemic uncertainty (\eg unfamiliar concepts in prompt) could lead to more divergent reconstructions.
\\
\textbf{LMD \cite{liu2023unsupervised}.} Another similar approach for image-space OOD detection, known as Learned Manifold Denoising (LMD), instead explores alternative transforms on $\b x^*_0$ to noising. The idea is to lift $\b x^*_0$ from its original data manifold and then map it back using the trained diffusion model. If the sample is in-distribution, the mapping process should return it close to its original position on the data manifold. Conversely, if it is OOD, the mapped position will likely diverge significantly from the original. The corruption process typically involves masking portions of the image, with various mask patterns applied. Similar to DDPM-OOD, similarity is measured between the original image and the reconstructed version after applying the diffusion process. Similar to before, since we lack a testimage, we measure the similarity between reconstructed images and \(\hat{\b{x}}_{0_i}\), the image generated directly from the prompt without perturbations.
\\
\textbf{Similarity Metric.} Various methods are used to evaluate similarity between images. Mean-squared error (MSE) is commonly applied between input and reconstruction, while some approaches favor the Learned Perceptual Image Patch Similarity (LPIPS)~\cite{zhang2018lpips} metric. The choice of similarity metric is a crucial component of these techniques.

\subsubsection{Test-Time Ensembling Approaches}
Test-time ensembling techniques take inspiration from approaches used in the large language model (LLM) community \cite{semantic_unc}, where multiple generations are produced to assess consistency. Given a prompt \(\vc\), we generate two images \(\{\hat{\b{x}}_{1},\hat{\b{x}}_{2}\}\) and measure the similarity between the two outputs. We denote this technique 2XDM. The underlying rationale is that if the model has high confidence, it will generate similar images. However, if uncertainty is high, it will generate substantially different images. We note that this is similar to the approach proposed in \cite{ensdiffdecu}, however, that approach uses an \textit{ensemble} of diffusion models, which is not practical in the setting where we want to quantify the uncertainty of a single pretrained T2I model.

A key question remains regarding how to find the best measure similarity, and we propose using the same techniques applied in the time-step-based methods. A significant advantage of test-time ensembling approaches is that they can be applied generically to all T2I models. In contrast, timestep-based approaches require a multi-step generation process, limiting the broadness of their applicability. We note that all above approaches require \textit{multiple} generations, incurring a notable computational overhead at inference time. For example, DDPM-OOD requires $~\sim 50\times$ the number of network inferences as a single generation (if the hyperparameters of the original paper are used) \cite{graham2023denoising}.


\subsection{Prompt-based Uncertainty  Estimation}\label{sec:punc}

In this section, we propose a new method, Prompt-based UNCertainty estimation for text-to-image generation (\method), which leverages Large Vision-Language Models (LVLM) in conjunction with T2I models. We aim to assess the uncertainty in generated images by analyzing the alignment between initial text prompts and reconstructed textual descriptions of the generated images.

\paragraph{Background on Large Vision-Language Model (LVLM)}

A LVLM integrates a large language model (LLM) with an image encoder, allowing for robust, multimodal understanding. Given a text prompt \(\vc\) and an image \(\vx\), a LVLM model, with parameters $\b \omega$, includes:

\begin{itemize}
    \item \textbf{Image Encoder:} \(f_{\b\omega}^{\text{img}}(\cdot)\) processes the input image \(\vx\) and produces an embedding: 
    $
    \vz^{\text{img}} = f_{\b\omega}^{\text{img}}(\vx)
    $

    \item \textbf{Text Processor with LLM:} \(f_{\b\omega}^{\text{txt}}(\cdot, \cdot)\) takes the prompt \(\vc\) and the image embedding \(\vz^{\text{img}}\) as inputs, generating a descriptive or answer-based response: 
    $
    \hat{\vc} = f_{\b\omega}^{\text{txt}}(\vc, \vz^{\text{img}})
    $
\end{itemize}
Thus, a LVLM system generates an interpretation \(\hat{\vc}\) that reflects the content of the image given an initial prompt.

\paragraph{\underline{P}rompt-Based \underline{UNC}ertainty Estimation for T2I Generation (\method)}

The previously introduced approaches to uncertainty estimation rely on measuring similarity between images. For a given text prompt \( \vc \), a T2I model generates an image \( \b{x} \sim p_\vtheta(\b{x} |  \vc) \). We reason that, even under low uncertainty, the model can produce images that vary in aspects such as color, texture, or object positioning, while still faithfully representing the semantics of prompt \( \vc \). As previously discussed in \cref{sec:uncertainty}, this may result in \textit{image-based} similarity metrics failing to capture uncertainty at a \textit{semantic} level, since large differences in the image space may not necessarily correspond to meaningful semantic differences. 

We propose that advanced LVLMs, such as Molmo~\cite{deitke2024molmo}, LLAMA 3~\cite{dubey2024llama}, and GPT-4~\cite{achiam2023gpt}, are a strong candidate for extracting the \textit{semantics} from an image, through describing an image via natural language. Building on this insight, our proposed method, \method, introduces a novel approach that bypasses direct image comparison and instead evaluates the alignment between generated images and the original prompts by calculating an uncertainty score in the more semantically meaningful \textit{text}-space. The steps for implementing \method~ are as follows:
\\
\textbf{Step 1: Initial Prompt and Image Generation}
Given a test prompt \(\vc^*\), we use the T2I model to sample an image \(\b{x}\):
\[
\b{x} \sim
p_\vtheta(\b{x} |  \vc^*) 
\]
\textbf{Step 2: Image Interpretation via LVLM}
With the generated image \(\vx\) and the initial prompt \(\vc^*\), the LVLM produces a new descriptive caption \(\hat{\vc}\):
\[
\hat{\vc} = f_{\b\omega}^{\text{txt}}(\vc^*, f_{\b\omega}^{\text{img}}(\vx))
\]
\textbf{Step 3: Uncertainty Score Calculation}
The core of \method~ lies in measuring the alignment between \(\vc\) and \(\hat{\vc}\) to assess the diffusion model's confidence. Specifically, the similarity score \(S(\vc, \hat{\vc})\) is computed as:
\[
S(\vc^*, \hat{\vc}) = \text{sim}(\vc^*, \hat{\vc})
\]
High similarity suggests low uncertainty, indicating the generated image closely matches the initial prompt. Conversely, a lower score reveals higher uncertainty, signifying potential divergence from the prompt's intended meaning.

Thus, \method~is able to predict semantic uncertainties by utilising the powerful image understanding ability of LVLMs and computing similarities in the \textit{text} space.
\\
\textbf{Prompt Similarity: Aleatoric and Epistemic via Precision and Recall.} In \cref{sec:uncertainty} we discussed the different types of semantic uncertainty for T2I generation. Here we will present a method for disentangling these uncertainties using the precision and recall of the caption given the prompt. Consider an abstract scenario where the prompt and the generated image each contain a discrete set of semantic concepts. We can then calculate the \textit{precision} and \textit{recall} of the image concepts with respect to the prompt:

\noindent\includegraphics[width=\columnwidth]{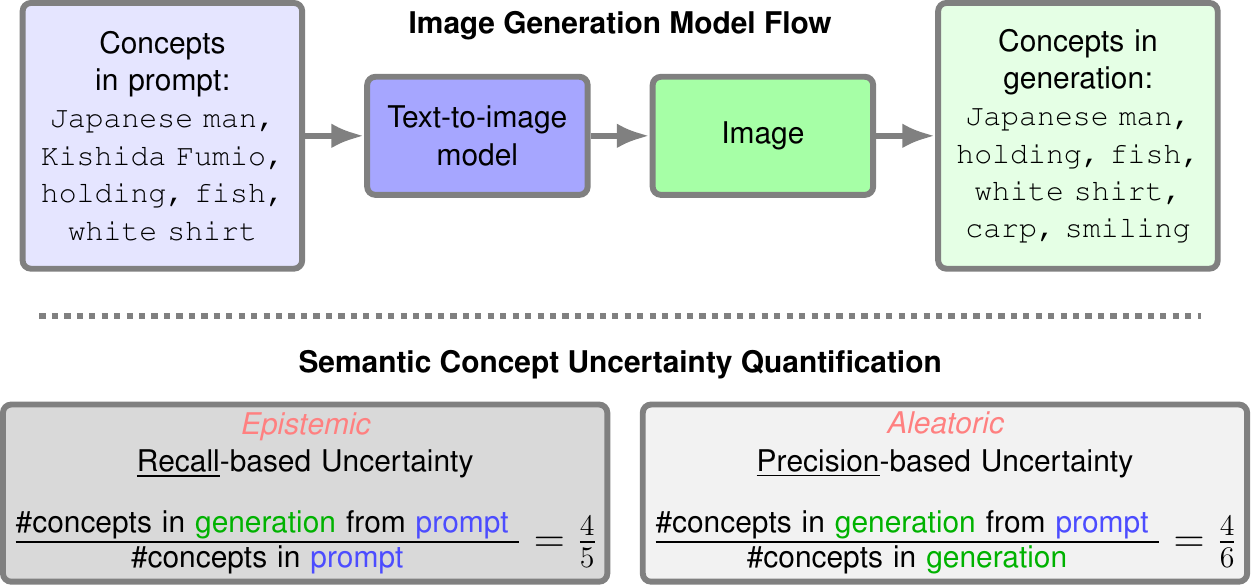}\\
Intuitively, a \textit{lack of knowledge} about concepts in the prompt (\textit{epistemic} uncertainty) will result in fewer concepts being preserved in the image, reducing \textit{recall}. On the other hand, a lack of specificity in the prompt (\textit{aleatoric} uncertainty) will result in additional concepts being generated, reducing \textit{precision}. Thus we can use this abstract form of semantic-concept-based recall to measure epistemic uncertainty, and use precision to measure aleatoric uncertainty.

Although we cannot directly measure concepts, we can approximate them using natural language. Moreover, existing similarity scores for text already have the above notions of precision and recall integrated within \cite{papineni2002bleu,lin2004rouge,zhang2019bertscore}, allowing \method~to disentangle uncertainty by comparing the caption of the generated image with the original prompt. We note that the \textit{image-based} approaches in \cref{sec:baseline_methods} are \textit{unable} to disentangle aleatoric from epistemic uncertainty. We focus on two different existing text-based similarity metrics, ROUGE \cite{lin2004rouge}, and BERTScore \cite{zhang2019bertscore}. The former calculates similarity (precision or recall) by directly matching (sequences of) words between two text sequences. The latter on the other hand measures similarity in the deep feature space of a text encoder. A more detailed discussion on how these scores are calculated can be found in \cref{sec:bertrouge}.

\section{Experiments \& Discussions\label{sec:xp_results}}

We test \method~ on Various task and various dataset and compared \method~ to DDPM-OOD \cite{graham2023denoising}, LMD \cite{liu2023unsupervised} and 2XDM. We first introduce on \cref{sec:Dataset} how we have constitute the dataset of prompts. Then on \cref{sec:Results} we discuss of different results on these dataset. Then \cref{sec:CopyrightPolitician} explains how we can use the uncertainty of text-to-image model to detect if a model has been trained with some concept related to some fictional character of politicians. \cref{sec:Bias} shows how uncertainty can help to detect bias of T2I models. 
To conduct our experiments and evaluate the various techniques for quantifying uncertainty, we utilized four T2I models: Stable Diffusion 1.5 (SDv1.5) \cite{rombach2022high}, SDXS \cite{song2024sdxs}, SDXL \cite{podellsdxl}, and PixArt-$\Sigma$ \cite{chen2024pixart}. These diverse T2I models allow us to test a range of architectures, each with unique characteristics and potential weaknesses. SDv1.5, being one of the earliest models, likely has a smaller and less diverse training dataset compared to the others. In contrast, SDXS generates in a single-step, different to the multi-step processes in other models.
\\
\textbf{Dataset of prompts}\label{sec:Dataset}
To conduct our experiments, we require both an in-distribution dataset of prompts and datasets with uncertainty. For the in-distribution prompts, we use the dataset proposed by \cite{ding2023gpt4image}, where high-quality descriptions of ImageNet \cite{deng2009imagenet} images are generated using GPT-4\cite{ding2023gpt4image}, referred to as \textit{Normal}. To simulate epistemic uncertainty, we create out-of-distribution (OOD) datasets using images from underrepresented domains like remote sensing, texture, and microscopic datasets, captioned with LLAVA Next\cite{li2024llava}, labeled as \textit{Remote Sensing}, \textit{Texture}, and \textit{Microscopic}. For aleatoric uncertainty, we generate two datasets: \textit{Vague}, containing 2,000 minimally descriptive prompts (e.g., "An image of ***"), and \textit{Corrupted}, derived from \textit{Normal} by adding grammatical errors (Level 1) and removing $50\%$ of words (Level 2). A detailed summary of these datasets is presented in \cref{tab:prompt_dataset} and \cref{sec:Dataset_Supp}.
In \cref{sec:Experimentalsettings}, we provide the details of the hyperparameters used in our experiments.

\begin{table}[ht]
\centering
\label{tab:prompt_datasets}
\resizebox{\columnwidth}{!}{%
\begin{tabular}{l|l|p{7cm}}
\hline
\textbf{Dataset Type} & \textbf{Dataset Name} & \textbf{Description} \\ \hline
\multirow{1}{*}{In-Distribution (ID)} 
& \textit{Normal} & 20k High-quality detailed prompts from ImageNet~\cite{deng2009imagenet} images using GPT-4\cite{ding2023gpt4image}.\\
\midrule
\multirow{3}{*}{Out-of-Distribution (OOD)} 
& \textit{Remote Sensing} & 20k Prompts from remote sensing images \cite{BigEarthNet}\\
& \textit{Texture} & 5k Prompts from texture datasets \\
& \textit{Microscopic} & 10k Prompts from microscopic images \cite{Microdata}\\ 
\midrule
\multirow{2}{*}{Aleatoric Uncertainty} 
& \textit{Vague} & 2k prompts with minimal context, e.g., ``An image of ***," where "***" is an ImageNet class name, creating ambiguous input. \\
& \textit{Adversarial} & 1k altered prompts from \textit{Normal} using UnlearnDiffAtk~\cite{zhang2024generatenotsafetydrivenunlearned}\\
& \textit{Corrupted} & 20k Altered prompts from \textit{Normal}:
   \begin{itemize}
      \item \textbf{Level 1}: Grammatical mistakes and spelling errors.
      \item \textbf{Level 2}: Random removal of $50\%$ of words in Level 1 prompts. 
   \end{itemize}
\\
   \bottomrule
\end{tabular}%
}
\caption{Summary of prompt datasets for unc. quantification.\label{tab:prompt_dataset}}
\end{table}

\subsection{Results for Uncertainty Estimation}\label{sec:Results}

\cref{tab:Epistemic} presents the results of \method~ when using Molmo as the LVLM to assess epistemic uncertainty in the T2I models. Our approach, \method, outperforms most state-of-the-art techniques on average. However, we observe more mixed results on the Texture dataset, likely due to uncertainty about the presence of texture images in the training data of the T2I models.

In \cref{tab:Aleatoric}, we show the results of \method~ under aleatoric uncertainty, again using Molmo as the LVLM. Our approach outperforms most state-of-the-art methods in this setting as well. Notably, while the ROUGE score tends to perform better for epistemic uncertainty, the optimal metric is less clear for aleatoric uncertainty. Here, the choice between ROUGE and BERTScore may depend on the specific application: because corruption introduces spelling and grammatical errors, BERTScore—which captures semantic similarity—may be more suitable. For additional results using different LVLMs, please refer to \cref{sec:ablation}. In \cref{sec:Qualitatifs}, we present qualitative results.

\subsection{Applications of uncertainty}\label{sec:Applications}

 Our study explores multiple applications of uncertainty quantification in T2I models, with detailed results provided in \cref{sec:CopyrightPolitician}. First, we investigate deepfake detection by examining the model's ability to generate recognizable images of prominent politicians (see \cref{sec:Deepfake}). Next, we address copyright concerns by evaluating whether T2I models can accurately replicate iconic, copyrighted characters such as Mickey Mouse and Darth Vader (see  \cref{sec:Copyright}). Finally, we analyze bias in T2I models by testing their capacity to generate individuals across different genders and races for specific job roles (see \cref{sec:Bias}). 

Using a LVLM, we evaluate the presence of specific individuals, copyrighted characters, or demographic concepts in the generated images, demonstrating how uncertainty estimation can aid in detecting potential misuses of generative models. These experiments illustrate the value of text-based uncertainty quantification in analyzing T2I model performance across ethically and legally sensitive applications, enabling assessments that are challenging to achieve with image-based generation alone.

\begin{table*}[t]
    \centering
    \resizebox{\textwidth}{!}
    {
\begin{tabular}{cl|ccc|ccc|ccc|ccc|ccc}
\toprule
\multicolumn{2}{l}{}                                         & \multicolumn{3}{c|}{SDXS}                                           & \multicolumn{3}{c|}{PixArt}                             & \multicolumn{3}{c|}{SDv1.5}                                            & \multicolumn{3}{c|}{SDXL}                               & \multicolumn{3}{c}{Average}                                                              \\
\multicolumn{1}{l}{}              &                          & auroc $\uparrow$             & aupr $\uparrow$              & fpr95$\downarrow$              & auroc $\uparrow$         & aupr $\uparrow$          & fpr95$\downarrow$          & auroc $\uparrow$         & aupr $\uparrow$          & fpr95$\downarrow$                         & auroc $\uparrow$         & aupr $\uparrow$          & fpr95$\downarrow$          & \multicolumn{1}{c}{auroc $\uparrow$} & \multicolumn{1}{c}{aupr $\uparrow$} & \multicolumn{1}{c}{fpr95$\downarrow$} \\
\midrule
                                  & DDPM-OOD mse             & \multicolumn{1}{l}{} & \multicolumn{1}{l}{} & \multicolumn{1}{l|}{} & 39.61\%          & 56.44\%          & 100.00\%         & 71.87\%          & 71.84\%          & 93.59\%                         & 87.39\%          & 83.91\%          & 51.10\%          & 66.29\%                      & 70.73\%                     & 81.56\%                     \\
                                  & DDPM-OOD lpips\_alex     & \multicolumn{1}{l}{} & \multicolumn{1}{l}{} & \multicolumn{1}{l|}{} & 24.24\%          & 32.53\%          & 100.00\%         & 22.62\%          & 33.04\%          & 99.69\%                         & 18.09\%          & 30.85\%          & 99.96\%          & 21.65\%                      & 32.14\%                     & 99.89\%                     \\
                                  & DDPM-OOD mse\_and\_lpips & \multicolumn{1}{l}{} & \multicolumn{1}{l}{} & \multicolumn{1}{l|}{} & 39.61\%          & 56.38\%          & 100.00\%         & 71.64\%          & 71.52\%          & 93.62\%                         & 87.25\%          & 83.67\%          & 51.43\%          & 66.17\%                      & 70.53\%                     & 81.68\%                     \\
                                  & LMD mse                  & \multicolumn{1}{l}{} & \multicolumn{1}{l}{} & \multicolumn{1}{l|}{} & 41.27\%          & 58.00\%          & 100.00\%         & 72.20\%          & 69.86\%          & 85.56\%                         & 55.06\%          & 48.72\%          & 89.80\%          & 56.17\%                      & 58.86\%                     & 91.79\%                     \\
                                  & LMD lpips\_alex          & \multicolumn{1}{l}{} & \multicolumn{1}{l}{} & \multicolumn{1}{l|}{} & 45.82\%          & 45.24\%          & 99.10\%          & 16.81\%          & 31.61\%          & 99.88\%                         & 5.75\%           & 28.12\%          & 100.00\%         & 22.79\%                      & 34.99\%                     & 99.66\%                     \\
                                  & 2XDM mse                 & 31.02\%              & 35.89\%              & 99.77\%              & 35.43\%          & 43.22\%          & 100.00\%         & 54.21\%          & 52.27\%          & 98.88\%                         & 65.66\%          & 52.85\%          & 76.71\%          & 46.58\%                      & 46.06\%                     & 93.84\%                     \\
                                  & 2XDM lpips\_alex         & 24.49\%              & 32.60\%              & 99.91\%              & 24.32\%          & 33.21\%          & 99.99\%          & 16.19\%          & 32.42\%          & 99.94\%                         & 30.58\%          & 36.99\%          & 98.25\%          & 23.90\%                      & 33.81\%                     & 99.52\%                     \\
                                  \cmidrule{2-17}
                                  & \method~ ROUGE  (1\_recall)  & 84.00\%              & 85.10\%              & 57.07\%              & 95.47\%          & 97.06\%          & 30.84\%          & 87.24\%          & 90.66\%          & 48.90\%                         & 83.56\%          & 84.31\%          & 52.78\%          & 87.57\%                      & 89.28\%                     & 47.40\%                     \\
                                  & \method~ ROUGE  (L\_recall)  & 87.64\%              & 89.56\%              & 55.28\%              & 96.96\%          & 98.02\%          & 16.85\%          & \textbf{90.20\%} & \textbf{93.40\%} & \textbf{43.56\%}                & \textbf{88.19\%} & \textbf{89.84\%} & \textbf{47.44\%} & \textbf{90.75\%}             & \textbf{92.71\%}            & \textbf{40.78\%}            \\

\multirow{-10}{*}{\rotatebox{90}{microscopic}}    & \method~ BERT (recall)      & \textbf{88.45\%}     & \textbf{89.71\%}     & \textbf{48.68\%}     & \textbf{97.83\%} & \textbf{98.41\%} & \textbf{10.32\%} & 87.75\%          & 91.70\%          & 51.03\%                         & 83.10\%          & 85.46\%          & 58.14\%          & 89.28\%                      & 91.32\%                     & 42.04\%                     \\
\midrule 
                                  & DDPM-OOD mse             & \multicolumn{1}{l}{} & \multicolumn{1}{l}{} & \multicolumn{1}{l|}{} & 79.96\%          & 85.31\%          & 99.28\%          & 74.16\%          & 71.64\%          & 71.18\%                         & 65.11\%          & 57.88\%          & 72.29\%          & 73.07\%                      & 71.61\%                     & 80.92\%                     \\
                                  & DDPM-OOD lpips\_alex     & \multicolumn{1}{l}{} & \multicolumn{1}{l}{} & \multicolumn{1}{l|}{} & 76.56\%          & 82.25\%          & 97.47\%          & \textbf{86.11\%} & \textbf{86.43\%} & 65.79\%                         & 85.10\%          & 86.12\%          & 58.96\%          & 82.59\%                      & 84.93\%                     & 74.07\%                     \\
                                  & DDPM-OOD mse\_and\_lpips & \multicolumn{1}{l}{} & \multicolumn{1}{l}{} & \multicolumn{1}{l|}{} & 80.00\%          & 85.37\%          & 99.28\%          & 74.36\%          & 71.89\%          & 71.00\%                         & 65.47\%          & 58.25\%          & 71.97\%          & 73.28\%                      & 71.84\%                     & 80.75\%                     \\
                                  & LMD mse                  & \multicolumn{1}{l}{} & \multicolumn{1}{l}{} & \multicolumn{1}{l|}{} & 83.95\%          & 89.53\%          & 99.05\%          & 75.76\%          & 71.88\%          & 66.83\%                         & 58.05\%          & 53.17\%          & 85.33\%          & 72.59\%                      & 71.53\%                     & 83.74\%                     \\
                                  & LMD lpips\_alex          & \multicolumn{1}{l}{} & \multicolumn{1}{l}{} & \multicolumn{1}{l|}{} & 90.65\%          & 93.11\%          & 74.49\%          & 78.93\%          & 79.31\%          & 81.05\%                         & 86.60\%          & 85.39\%          & 49.19\%          & 85.39\%                      & 85.94\%                     & 68.24\%                     \\
                                  & 2XDM mse                 & 60.13\%              & 56.57\%              & 86.40\%              & 56.52\%          & 54.87\%          & 99.26\%          & 28.62\%          & 39.11\%          & 98.44\%                         & 9.28\%           & 31.95\%          & 98.99\%          & 38.64\%                      & 45.62\%                     & 95.77\%                     \\
                                  & 2XDM lpips\_alex         & 53.88\%              & 59.21\%              & 98.23\%              & 54.43\%          & 57.99\%          & 98.78\%          & 30.66\%          & 38.80\%          & 98.76\%                         & 23.54\%          & 35.35\%          & 96.23\%          & 40.63\%                      & 47.83\%                     & 98.00\%                     \\
                                  \cmidrule{2-17}
                                  & \method~ ROUGE  (1\_recall)  & 85.59\%              & 86.59\%              & 63.87\%              & 94.14\%          & 95.97\%          & 49.29\%          & 73.24\%          & 72.93\%          & 79.56\%                         & 73.07\%          & 74.54\%          & 83.48\%          & 81.51\%                      & 82.51\%                     & 69.05\%                     \\
                                  & \method~ ROUGE  (L\_recall)  & \textbf{89.85\%}     & \textbf{90.63\%}     & \textbf{51.96\%}     & \textbf{97.01\%} & \textbf{97.94\%} & \textbf{16.08\%} & 83.63\%          & 84.65\%          & \textbf{65.35\%}                & \textbf{85.68\%} & \textbf{87.09\%} & \textbf{64.00\%} & \textbf{89.04\%}             & \textbf{90.08\%}            & \textbf{49.34\%}            \\
                 
\multirow{-10}{*}{\rotatebox{90}{remote sensing}} & \method~ BERT (recall)      & 80.50\%              & 82.56\%              & 75.19\%              & 91.93\%          & 93.08\%          & 43.30\%          & 61.27\%          & 64.74\%          & 90.08\%                         & 60.71\%          & 65.32\%          & 90.94\%          & 73.60\%                      & 76.42\%                     & 74.88\%                     \\
\midrule      
                                  & DDPM-OOD mse             & \multicolumn{1}{l}{} & \multicolumn{1}{l}{} & \multicolumn{1}{l|}{} & 73.27\%          & 62.14\%          & 89.25\%          & 70.33\%          & 56.23\%          & 99.80\%                         & 74.33\%          & 52.19\%          & \textbf{84.62\%}          & 72.64\%                      & 56.86\%                     & 91.22\%                     \\
                                  & DDPM-OOD lpips\_alex     & \multicolumn{1}{l}{} & \multicolumn{1}{l}{} & \multicolumn{1}{l|}{} & 43.94\%          & 27.22\%          & 97.83\%          & 34.70\%          & 19.45\%          & 99.62\%                         & 34.37\%          & 22.32\%          & 99.97\%          & 37.67\%                      & 22.99\%                     & 99.14\%                     \\
                                  & DDPM-OOD mse\_and\_lpips & \multicolumn{1}{l}{} & \multicolumn{1}{l}{} & \multicolumn{1}{l|}{} & 73.22\%          & 62.02\%          & 89.24\%          & 70.21\%          & 55.97\%          & 99.80\%                         & \textbf{74.24\%} & 51.97\%          & 84.71\%          & \textbf{72.56\%}             & 56.65\%                     & 91.25\%                     \\
                                  & LMD mse                  & \multicolumn{1}{l}{} & \multicolumn{1}{l}{} & \multicolumn{1}{l|}{} & 81.09\%          & 71.12\%          & 80.52\%          & \textbf{71.44\%} & 51.24\%          & 99.81\%                         & 56.52\%          & 31.20\%          & 95.00\%          & 69.68\%                      & 51.19\%                     & 91.78\%                     \\
                                  & LMD lpips\_alex          & \multicolumn{1}{l}{} & \multicolumn{1}{l}{} & \multicolumn{1}{l|}{} & 63.40\%          & 50.47\%          & 97.52\%          & 26.74\%          & 16.06\%          & 99.89\%                         & 35.66\%          & 23.22\%          & 99.95\%          & 41.93\%                      & 29.92\%                     & 99.12\%                     \\
                                  & 2XDM mse                 & \textbf{57.17}\%              & 27.12\%              & 97.11\%              & 54.19\%          & 24.78\%          & 92.23\%          & 46.89\%          & 20.82\%          & 99.83\%                         & 51.23\%          & 21.29\%          & 96.95\%          & 52.37\%                      & 23.50\%                     & 96.53\%                     \\
                                  & 2XDM lpips\_alex         & 32.80\%              & 20.84\%              & 99.98\%              & 37.21\%          & 23.36\%          & 99.37\%          & 29.58\%          & 24.27\%          & 99.93\%                         & 28.87\%          & 16.49\%          & 98.88\%          & 32.11\%                      & 21.24\%                     & 99.54\%                     \\
                                  \cmidrule{2-17}
    & \method~ ROUGE  (1\_recall)  & 40.77\%              & 72.88\%              & 95.54\%              & 83.26\%          & 94.80\%          & 73.48\%          & 46.61\%          & 75.64\%          & \textbf{90.87}\%                         & 49.16\%          & 78.04\%          & 93.57\%          & 54.95\%                      & 80.34\%                     & \textbf{88.36\%}                     \\
    & \method~ ROUGE  (L\_recall)  & 45.64\%              & 76.90\%              & 95.99\%              & 84.10\%          & 95.13\%          & 74.21\%          & 47.60\%          & 77.70\%          & 93.14\%                         & 50.71\%          & \textbf{80.05\%}          & 94.35\%          & 57.01\%                      & 82.44\%                     & 89.42\%                     \\
\multirow{-10}{*}{\rotatebox{90}{Texture}}        & \method~ BERT (recall)      & 49.11\%              & \textbf{79.11\%}              & \textbf{94.65\%}              & \textbf{84.40\%} & \textbf{95.14\%} & \textbf{69.99\%} & 47.60\%          & \textbf{78.80\%}          & 93.85\%                         & 46.72\%          & 79.20\%          & 96.87\%          & 56.96\%                      & \textbf{83.06}\%                     & 88.84\%\\
    \bottomrule
\end{tabular}
    }
    \caption{\textbf{Performance comparison of \method~ applied to different Out of Distribution concepts and different text-to-image models}}
    \label{tab:Epistemic}
\end{table*}

\begin{table*}[tbhp!]
\centering
    \resizebox{\textwidth}{!}
    {
\begin{tabular}{cl|ccc|ccc|ccc|ccc|ccc}
\toprule
\multicolumn{2}{l}{}                                         & \multicolumn{3}{c|}{SDXS}                                           & \multicolumn{3}{c|}{PixArt}                             & \multicolumn{3}{c|}{SDv1.5}                                            & \multicolumn{3}{c|}{SDXL}                               & \multicolumn{3}{c}{Average}                                                              \\
\multicolumn{1}{l}{}              &                          & auroc $\uparrow$             & aupr $\uparrow$              & fpr95$\downarrow$              & auroc $\uparrow$         & aupr $\uparrow$          & fpr95$\downarrow$          & auroc $\uparrow$         & aupr $\uparrow$          & fpr95$\downarrow$                         & auroc $\uparrow$         & aupr $\uparrow$          & fpr95$\downarrow$          & \multicolumn{1}{c}{auroc $\uparrow$} & \multicolumn{1}{c}{aupr $\uparrow$} & \multicolumn{1}{c}{fpr95$\downarrow$} \\
\midrule

\multirow{14}{*}{\rotatebox{90}{Vague}} & DDPM-OOD mse & \multicolumn{1}{l}{} & \multicolumn{1}{l}{} & \multicolumn{1}{l|}{} & 31.44\% & 6.07\% & 96.45\% & 50.07\% & 9.29\% & 96.56\% & 40.63\% & 5.36\% & 97.34\% & \multicolumn{1}{r}{40.71\%} & \multicolumn{1}{r}{6.91\%} & \multicolumn{1}{r}{96.78\%} \\
 & DDPM-OOD lpips\_alex & \multicolumn{1}{l}{} & \multicolumn{1}{l}{} & \multicolumn{1}{l|}{} & 33.60\% & 6.30\% & 97.45\% & 48.08\% & 10.09\% & 97.43\% & 46.95\% & 6.27\% & 98.53\% & \multicolumn{1}{r}{42.88\%} & \multicolumn{1}{r}{7.55\%} & \multicolumn{1}{r}{97.80\%} \\
 & DDPM-OOD mse\_and\_lpips & \multicolumn{1}{l}{} & \multicolumn{1}{l}{} & \multicolumn{1}{l|}{} & 31.41\% & 6.06\% & 96.45\% & 50.06\% & 9.30\% & 96.52\% & 40.59\% & 5.36\% & 97.37\% & \multicolumn{1}{r}{40.69\%} & \multicolumn{1}{r}{6.91\%} & \multicolumn{1}{r}{96.78\%} \\
 & LMD mse & \multicolumn{1}{l}{} & \multicolumn{1}{l}{} & \multicolumn{1}{l|}{} & 31.59\% & 6.09\% & 97.83\% & 48.32\% & 8.84\% & 95.78\% & 41.27\% & 5.72\% & 97.79\% & \multicolumn{1}{r}{40.39\%} & \multicolumn{1}{r}{6.88\%} & \multicolumn{1}{r}{97.13\%} \\
 & LMD lpips\_alex & \multicolumn{1}{l}{} & \multicolumn{1}{l}{} & \multicolumn{1}{l|}{} & 39.26\% & 6.91\% & 96.35\% & 47.62\% & 9.54\% & 97.70\% & 40.77\% & 5.70\% & 97.95\% & \multicolumn{1}{r}{42.55\%} & \multicolumn{1}{r}{7.38\%} & \multicolumn{1}{r}{97.33\%} \\
 & 2XDM mse & 82.69\% & 55.64\% & 88.32\% & 44.25\% & 8.77\% & 94.35\% & 51.42\% & 9.89\% & 96.01\% & 56.58\% & 9.46\% & 95.69\% & \multicolumn{1}{r}{58.74\%} & \multicolumn{1}{r}{20.94\%} & \multicolumn{1}{r}{93.59\%} \\
 & 2XDM lpips\_alex & 51.28\% & 10.10\% & 94.35\% & 43.82\% & 8.57\% & 96.89\% & 50.97\% & 10.09\% & 95.96\% & 57.83\% & 10.01\% & 94.37\% & \multicolumn{1}{r}{50.98\%} & \multicolumn{1}{r}{9.69\%} & \multicolumn{1}{r}{95.39\%} \\
 \cmidrule{2-17}
 & \method~ ROUGE (1\_precision) & \textbf{99.99\%} & \textbf{100.00\%} & 0.00\% & \textbf{100.00\%} & \textbf{100.00\%} & \textbf{0.00\%} & \textbf{99.99\%} & \textbf{100.00\%} & \textbf{0.00\%} & \textbf{99.99\%} & \textbf{100.00\%} & 0.00\% & \multicolumn{1}{r}{\textbf{99.99\%}} & \multicolumn{1}{r}{\textbf{100.00\%}} & \multicolumn{1}{r}{0.00\%} \\
 & \method~ ROUGE (L\_precision) & \textbf{99.99\%} & \textbf{100.00\%} & 0.00\% & \textbf{100.00\%} & \textbf{100.00\%} & \textbf{0.00\%} & \textbf{99.99\%} & \textbf{100.00\%} & \textbf{0.00\%} & \textbf{100.00\%} & \textbf{100.00\%} & 0.00\% & \multicolumn{1}{r}{\textbf{100.00\%}} & \multicolumn{1}{r}{\textbf{100.00\%}} & \multicolumn{1}{r}{0.00\%} \\
 & \method~  BERT (precision) & 60.63\% & 92.24\% & 82.85\% & 67.02\% & 94.54\% & 83.10\% & 37.35\% & 86.51\% & 92.70\% & 70.85\% & 95.13\% & 78.00\% & \multicolumn{1}{r}{58.96\%} & \multicolumn{1}{r}{92.11\%} & \multicolumn{1}{r}{84.16\%} \\
 \midrule 
\multirow{9}{*}{\rotatebox{90}{Adversarial}} & DDPM-OOD mse & \multicolumn{1}{l}{} & \multicolumn{1}{l}{} & \multicolumn{1}{l|}{} & \textit{39.56\%} & \textit{1.78\%} & \textit{95.72\%} & \textit{50.86\%} & \textit{2.53\%} & \textit{95.92\%} & \textit{38.98\%} & \textit{1.10\%} & \textit{98.38\%} & 43.14\% & 1.81\% & 96.67\% \\
 & DDPM-OOD lpips\_alex & \multicolumn{1}{l}{} & \multicolumn{1}{l}{} & \multicolumn{1}{l|}{} & \textit{39.59\%} & \textit{1.81\%} & \textit{97.16\%} & \textit{53.65\%} & \textit{3.40\%} & \textit{96.20\%} & \textit{50.39\%} & \textit{1.42\%} & \textit{97.26\%} & 47.88\% & 2.21\% & 96.87\% \\
 & DDPM-OOD mse\_and\_lpips & \multicolumn{1}{l}{} & \multicolumn{1}{l}{} & \multicolumn{1}{l|}{} & \textit{39.54\%} & \textit{1.78\%} & \textit{95.72\%} & \textit{50.89\%} & \textit{2.54\%} & \textit{95.91\%} & \textit{38.97\%} & \textit{1.10\%} & \textit{98.38\%} & 43.13\% & 1.81\% & 96.67\% \\
 & LMD mse & \multicolumn{1}{l}{} & \multicolumn{1}{l}{} & \multicolumn{1}{l|}{} & \textit{40.74\%} & \textit{1.83\%} & \textit{97.41\%} & \textit{46.95\%} & \textit{2.27\%} & \textit{96.35\%} & \textit{37.90\%} & \textit{1.09\%} & \textit{98.80\%} & 41.87\% & 1.73\% & 97.52\% \\
 & LMD lpips\_alex & \multicolumn{1}{l}{} & \multicolumn{1}{l}{} & \multicolumn{1}{l|}{} & \textit{43.71\%} & \textit{1.95\%} & \textit{96.24\%} & \textit{53.29\%} & \textit{2.89\%} & \textit{94.43\%} & \textit{41.64\%} & \textit{1.17\%} & \textit{97.48\%} & 46.21\% & 2.00\% & 96.05\% \\
 & 2XDM mse & \textit{58.34\%} & \textit{12.65\%} & \textit{98.86\%} & \textit{49.40\%} & \textit{2.45\%} & \textit{93.75\%} & \textit{53.12\%} & \textit{2.75\%} & \textit{96.46\%} & \textit{57.46\%} & \textit{2.09\%} & \textit{92.32\%} & 54.58\% & 4.99\% & 95.35\% \\
 & 2XDM lpips\_alex & \textit{36.94\%} & \textit{1.94\%} & \textit{98.92\%} & \textit{47.95\%} & \textit{2.54\%} & \textit{96.03\%} & \textit{53.04\%} & \textit{2.83\%} & \textit{95.69\%} & \textit{55.19\%} & \textit{1.96\%} & \textit{96.20\%} & 48.28\% & 2.32\% & 96.71\% \\
 \cmidrule{2-17}
 & \method~ ROUGE (1\_precision) & \textit{64.14\%} & \textit{97.77\%} & \textit{48.80\%} & \textit{76.32\%} & \textit{98.90\%} & \textit{48.00\%} & \textit{63.59\%} & \textit{97.73\%} & \textit{49.50\%} & \textit{62.97\%} & \textit{97.64\%} & \textit{49.20\%} & 66.76\% & 98.01\% & 48.87\% \\
 & \method~ ROUGE (L\_precision) & \textit{63.20\%} & \textit{97.75\%} & \textit{49.00\%} & \textit{76.26\%} & \textit{98.90\%} & \textit{48.40\%} & \textit{64.09\%} & \textit{97.90\%} & \textit{\textbf{49.70\%}} & \textit{63.76\%} & \textit{97.87\%} & \textit{49.60\%} & 66.83\% & 98.11\% & 49.17\% \\
 & \method~  BERT (precision) & \textit{73.98\%} & \textit{98.72\%} & \textit{78.31\%} & \textit{72.95\%} & \textit{98.93\%} & \textit{86.40\%} & \textit{45.09\%} & \textit{96.93\%} & \textit{94.18\%} & \textit{72.69\%} & \textit{98.80\%} & \textit{81.40\%} & 66.18\% & 98.34\% & 85.07\% \\
 \midrule 
\multirow{9}{*}{\rotatebox{90}{Corrupt lvl1}} & DDPM-OOD mse & \multicolumn{1}{l}{} & \multicolumn{1}{l}{} & \multicolumn{1}{l|}{} & 48.89\% & 44.83\% & 87.43\% & 53.15\% & 54.39\% & 95.48\% & 41.01\% & 29.53\% & 96.34\% & 47.68\% & 42.92\% & 93.08\% \\
 & DDPM-OOD lpips\_alex & \multicolumn{1}{l}{} & \multicolumn{1}{l}{} & \multicolumn{1}{l|}{} & 47.05\% & 45.74\% & 89.31\% & \textbf{58.30\%} & \textbf{58.59\%} & 93.56\% & 53.71\% & 37.37\% & 92.42\% & 53.02\% & 47.23\% & 91.76\% \\
 & DDPM-OOD mse\_and\_lpips & \multicolumn{1}{l}{} & \multicolumn{1}{l}{} & \multicolumn{1}{l|}{} & 48.87\% & 44.82\% & 87.43\% & 53.20\% & 54.43\% & 95.48\% & 41.02\% & 29.54\% & 96.32\% & 47.70\% & 42.93\% & 93.08\% \\
 & LMD mse & \multicolumn{1}{l}{} & \multicolumn{1}{l}{} & \multicolumn{1}{l|}{} & 51.34\% & 45.94\% & 86.93\% & 49.03\% & 51.08\% & 96.32\% & 34.80\% & 27.24\% & 98.14\% & 45.06\% & 41.42\% & 93.80\% \\
 & LMD lpips\_alex & \multicolumn{1}{l}{} & \multicolumn{1}{l}{} & \multicolumn{1}{l|}{} & 51.23\% & 47.48\% & 89.99\% & 57.53\% & 56.49\% & 90.92\% & 40.94\% & 30.30\% & 96.32\% & 49.90\% & 44.76\% & 92.41\% \\
 & 2XDM mse & 33.98\% & 44.40\% & 99.33\% & 57.04\% & 54.03\% & 87.71\% & 53.23\% & 52.17\% & 95.62\% & 52.80\% & 37.94\% & 94.55\% & 49.26\% & 47.14\% & 94.30\% \\
 & 2XDM lpips\_alex & 25.48\% & 36.16\% & 99.29\% & 50.43\% & 49.86\% & 90.10\% & 54.97\% & 54.29\% & 95.69\% & 54.73\% & 38.27\% & 93.29\% & 46.40\% & 44.65\% & 94.59\% \\
 \cmidrule{2-17}
 & \method~ ROUGE (1\_precision) & 28.75\% & 38.34\% & 98.77\% & 54.58\% & 55.29\% & 96.02\% & 27.69\% & 37.57\% & 98.78\% & 27.45\% & 37.27\% & 98.54\% & 34.62\% & 42.12\% & 98.03\% \\
 & \method~ ROUGE (L\_precision) & 27.29\% & 38.11\% & 99.15\% & 54.32\% & 55.59\% & 96.16\% & 27.30\% & 38.07\% & 99.45\% & 27.35\% & 37.89\% & 99.37\% & 34.07\% & 42.42\% & 98.53\% \\
 & \method~  BERT (precision) & \textbf{85.85\%} & \textbf{84.45\%} & \textbf{73.94\%} & \textbf{79.43\%} & \textbf{78.83\%} & 87.21\% & \textbf{54.97\%} & \textbf{53.73\%} & 94.51\% & \textbf{75.76\%} & \textbf{72.91\%} & 84.65\% & \textbf{74.00\%} & \textbf{72.48\%} & \textbf{85.08\%} \\
 \midrule 
\multirow{9}{*}{\rotatebox{90}{Corrupt lvl2}} & DDPM-OOD mse & \multicolumn{1}{l}{} & \multicolumn{1}{l}{} & \multicolumn{1}{l|}{} & \textit{35.94\%} & \textit{39.33\%} & \textit{92.36\%} & \textit{54.47\%} & \textit{56.66\%} & \textit{96.21\%} & \textit{37.25\%} & \textit{28.45\%} & \textit{97.12\%} & 42.55\% & 41.48\% & 95.23\% \\
 & DDPM-OOD lpips\_alex & \multicolumn{1}{l}{} & \multicolumn{1}{l}{} & \multicolumn{1}{l|}{} & \textit{43.23\%} & \textit{43.94\%} & \textit{91.23\%} & \textit{59.46\%} & \textit{60.14\%} & \textit{93.71\%} & \textit{48.79\%} & \textit{34.44\%} & \textit{95.74\%} & 50.49\% & 46.17\% & 93.56\% \\
 & LMD mse & \multicolumn{1}{l}{} & \multicolumn{1}{l}{} & \multicolumn{1}{l|}{} & \textit{39.88\%} & \textit{41.02\%} & \textit{92.72\%} & \textit{50.48\%} & \textit{52.68\%} & \textit{96.59\%} & \textit{40.28\%} & \textit{31.01\%} & \textit{97.97\%} & 43.55\% & 41.57\% & 95.76\% \\
 & LMD lpips\_alex & \multicolumn{1}{l}{} & \multicolumn{1}{l}{} & \multicolumn{1}{l|}{} & \textit{46.90\%} & \textit{45.65\%} & \textit{92.04\%} & \textit{\textbf{59.94\%}} & \textit{\textbf{58.90\%}} & \textit{\textbf{90.54\%}} & \textit{40.88\%} & \textit{30.42\%} & \textit{96.35\%} & 49.24\% & 44.99\% & 92.98\% \\
 & 2XDM mse & \textit{55.88\%} & \textit{59.72\%} & \textit{95.30\%} & \textit{51.98\%} & \textit{52.68\%} & \textit{90.13\%} & \textit{53.08\%} & \textit{53.10\%} & \textit{96.74\%} & \textit{55.06\%} & \textit{40.98\%} & \textit{94.80\%} & 54.00\% & 51.62\% & 94.24\% \\
 & 2XDM lpips\_alex & \textit{58.15\%} & \textit{58.50\%} & \textit{93.98\%} & \textit{52.08\%} & \textit{53.36\%} & \textit{90.19\%} & \textit{53.57\%} & \textit{54.17\%} & \textit{96.34\%} & \textit{60.24\%} & \textit{44.23\%} & \textit{91.78\%} & 56.01\% & 52.57\% & 93.07\% \\
 \cmidrule{2-17}
 & \method~ ROUGE (1\_precision) & \textit{28.76\%} & \textit{38.10\%} & \textit{98.59\%} & \textit{54.77\%} & \textit{55.49\%} & \textit{96.02\%} & \textit{27.66\%} & \textit{37.42\%} & \textit{98.75\%} & \textit{27.39\%} & \textit{37.23\%} & \textit{98.53\%} & 34.64\% & 42.06\% & 97.97\% \\
 & \method~ ROUGE (L\_precision) & \textit{27.33\%} & \textit{37.87\%} & \textit{99.16\%} & \textit{54.51\%} & \textit{55.87\%} & \textit{96.11\%} & \textit{27.31\%} & \textit{37.92\%} & \textit{99.35\%} & \textit{27.33\%} & \textit{37.83\%} & \textit{99.36\%} & 34.12\% & 42.37\% & 98.50\% \\
 & \method~  BERT (precision) & \textit{\textbf{85.96\%}} & \textit{\textbf{84.23\%}} & \textit{\textbf{73.54\%}} & \textit{\textbf{79.47\%}} & \textit{\textbf{78.80\%}} & \textit{\textbf{87.05\%}} & \textit{54.78\%} & \textit{53.51\%} & \textit{94.79\%} & \textit{\textbf{76.04\%}} & \textit{\textbf{73.44\%}} & \textit{\textbf{84.58\%}} & \textbf{74.06\%} & \textbf{72.49\%} & \textbf{84.99\%} \\ 
     \bottomrule

\end{tabular}
}
    \caption{
        \textbf{Performance comparison all applications (\textit{Vague}, \textit{Adversarial}, \textit{Corrupt Lvl1},\textit{ Corrup Lvl2})} \label{tab:Aleatoric}}
\end{table*}
\section{Conclusion}

In this work, we set out to be first to investigate uncertainty estimation T2I generation, where the uncertainty is with respect to the text prompt.  We both adapt existing image-space uncertainty approaches, as well as present Prompt-based UNCertainty Estimation for T2I models (\method), the first approach designed specifically to quantify uncertainty in T2I generation. By leveraging the interpretive power of Large Vision-Language Models to extract semantics from a generated image, \method~ addresses the challenges associated with measuring multimodal uncertainty, enabling a deeper understanding of model behavior with respect to prompt-based conditioning. Our results validate that \method~ provides reliable, fine-grained assessments of aleatoric and epistemic uncertainties, outperforming existing methods that work for image generation.

\method~ opens avenues for new applications, such as improving model robustness in biased or OOD scenarios and safeguarding against unauthorized content generation, including deepfakes and copyrighted characters. Our experiments show that diffusion models, while powerful, still exhibit significant variances in their ability to adhere to prompt details, particularly with ambiguous, corrupt prompts or OOD prompts. By identifying such uncertainties, \method~ not only supports the development of more trustworthy generative AI but also aids in aligning these models with ethical and regulatory standards.

Our contributions also include a dataset specifically tailored to uncertainty quantification in text-to-image models, which we hope will serve as a benchmark for future research. Ultimately, this work highlights the potential of prompt-based uncertainty estimation as a critical tool for enhancing the reliability and safety of generative models in real-world applications.

\textbf{Limitations.} While \method~ demonstrates promising results in quantifying uncertainty for text-to-image generation, several limitations remain. First, our approach relies heavily on the interpretive abilities of LVLMs, which, despite their effectiveness, may introduce biases or inaccuracies inherited from their training data. Second, our analysis focuses primarily on English-language prompts, limiting our method’s applicability to multilingual or culturally diverse datasets. 

\paragraph{Acknowledgments.}
 This work was performed using HPC resources from GENCI-IDRIS (Grant 2024-[AD011011970R4]).
\clearpage
{
    \small
    \bibliographystyle{ieeenat_fullname}
    \bibliography{main}
}

\appendix
\clearpage
\setcounter{page}{1}
\maketitlesupplementary

\section{Prompt similarity score}\label{sec:bertrouge}

 ROUGE \cite{lin2004rouge} and BERTScore \cite{zhang2019bertscore} are conceptually similar scores that measure how well a candidate text sequence (caption in our case) matches a reference text sequence (prompt). ROUGE operates at the text/token level whilst BERTScore is calculated on the deep embeddings of a pre-trained text encoder. Concretely they are calculated as follows:

\subsubsection*{BERTScore}

\begin{equation}
\text{Precision} = \frac{1}{|C|} \sum_{c \in C} \max_{r \in R} \text{cos}(c, r)
\end{equation}
\begin{equation}
\text{Recall} = \frac{1}{|R|} \sum_{r \in R} \max_{c \in C} \text{cos}(r, c)
\end{equation}
Where \( C \) and \( R \) are the sets of token embeddings for the candidate and reference, respectively. \( \text{cos}(x, y) \) is the cosine similarity between two embeddings.

\subsubsection*{ROUGE-n}

\begin{equation}\label{eq:rougen-p}
\text{Precision} = \frac{\sum_{g \in G(C)} \min(\text{count}(g, C), \text{count}(g, R))}{\sum_{g \in G(C)} \text{count}(g, C)}
\end{equation}

\begin{equation}\label{eq:rougen-r}
\text{Recall} =\frac{\sum_{g \in G(C)} \min(\text{count}(g, C), \text{count}(g, R))}{\sum_{g \in G(R)} \text{count}(g, R)}
\end{equation}
Where \( G(C) \) and \( G(R) \) are the sets of n-grams for the candidate and reference, respectively. \( \text{count}(g, S) \) is the count of n-gram \( g \) in sequence \( S \).

\subsubsection*{ROUGE-L}

\begin{equation}
\text{Precision} = \frac{\text{LCS}(C, R)}{|C|}
\end{equation}

\begin{equation}
\text{Recall} = \frac{\text{LCS}(C, R)}{|R|}
\end{equation}
Where \( \text{LCS}(C, R) \) is the length of the longest common subsequence between \( C \) and \( R \). \( |C| \) and \( |R| \) are the lengths of the candidate and reference sequences.

Comparing to the ideas of semantic-concept-based recall and precision presented in \cref{sec:punc}, we can see how the above formulae approximate concepts with \eg n-grams, BERT embeddings. \cref{eq:rougen-p,eq:rougen-r} directly substitute semantic concepts with n-grams for example, allowing for the approximate quantification of precision and recall based semantic uncertainty.

\section{Experimental settings}\label{sec:Experimentalsettings}
In this work, we evaluate the performance and capabilities of \method{} under various experimental conditions, testing several text-to-image (T2I) models: Stable Diffusion 1.5 (SDv1.5) \cite{rombach2022high}, SDXL \cite{podellsdxl}, PixArt-$\Sigma$ \cite{chen2024pixart}, and SDXS \cite{song2024sdxs}. All models are sourced from the Hugging Face repository and used with their respective default configurations, as summarized in \cref{tab:t2i_params}.

For inference, SDXS utilizes a single-step generation process, whereas the other models perform 20 inference steps. Regarding the guidance scale, SDXS applies no guidance, PixArt-$\Sigma$ uses a scale of 4.5, and both SDv1.5 and SDXL employ a scale of 7.5.

To evaluate the influence of various Large Vision-Language Models (LVLMs) and their caption generation capabilities on the performance of \method{}, we conduct experiments using three models: LLAMA3.2 Vision~\cite{dubey2024llama}, LLava-Next~\cite{li2024llava}, and Molmo~\cite{deitke2024molmo}. All models are implemented with their default configurations as provided in the Hugging Face library, and their specifications are summarized in \cref{tab:lvlm_params}.

Each model is sourced directly from the Hugging Face repository and evaluated under its standard settings. For the experiments, we use the maximum token length permissible for each model in relation to the given prompts. This ensures a consistent and comprehensive comparison of their captioning capabilities across the tested scenarios. Note that for SDXL, SDv1.5, and SDXS, the maximum token length is limited to 77 tokens, while for PixArt-$\Sigma$, a maximum of 300 tokens is permitted.

\begin{table}[h!]
\centering
\resizebox{\columnwidth}{!}
    {
\begin{tabular}{|c|c|c|c|}
\hline
T2I Model              & Inference Steps & Guidance Scale & Version  \\
\hline
SD 1.5 & 20              & 7.5  & \href{https://huggingface.co/sd-legacy/stable-diffusion-v1-5}{sd-legacy/stable-diffusion-v1-5}        \\
SDXL            & 20              & 7.5     &\href{https://huggingface.co/stabilityai/stable-diffusion-xl-base-1.0}{stabilityai/stable-diffusion-xl-base-1.0}    \\
PixArt-$\Sigma$    & 20              & 4.5   & \href{https://huggingface.co/PixArt-alpha/PixArt-Sigma}{PixArt-alpha/PixArt-Sigma}      \\
SDXS              & 1               & None        & \href{https://huggingface.co/IDKiro/sdxs-512-0.9}{IDKiro/sdxs-512-0.9}    \\
\hline
\end{tabular} 
}
\caption{Summary of model parameters used in the experiments. All models are sourced from the Hugging Face repository.}
\label{tab:t2i_params}
\end{table}

\begin{table}[h!]
\centering
\resizebox{\columnwidth}{!}
    {
\begin{tabular}{|c|c|c|c|}
\hline
LVLM              & Instruct &\# Params & Version  \\
\hline
LLama3.2- Vision & Yes   & 11B           & \href{https://huggingface.co/meta-llama/Llama-3.2-11B-Vision-Instruct}{meta-llama/Llama-3.2-11B-Vision-Instruct}         \\
Llava-Next    & Yes        & 7B      & \href{https://huggingface.co/llava-hf/llava-v1.6-mistral-7b-hf}{llava-hf/llava-v1.6-mistral-7b-hf}         \\
Molmo    & No   & 7B              & \href{https://huggingface.co/allenai/Molmo-7B-O-0924}{allenai/Molmo-7B-O-0924}         \\
\hline
\end{tabular}
}
\caption{Summary of model parameters used in the experiments. All models are sourced from the Hugging Face repository.}
\label{tab:lvlm_params}
\end{table}

\section{Ablation Results}\label{sec:ablation}
This section presents additional experiments as part of our ablation study, focusing on the model choices for \method{}, particularly the selection of Large Vision-Language Models (LVLMs) that form the core of \method{}. The results of these experiments are provided in Table~\ref{tab:Epistemic_supp} for the epistemic datasets and Table~\ref{tab:Aleatoric_supp} for the aleatoric datasets.

\begin{table*}[t]
    \centering
    \resizebox{\textwidth}{!}
    {
\begin{tabular}{cl|ccc|ccc|ccc|ccc|ccc}
\toprule
\multicolumn{2}{l}{}                                         & \multicolumn{3}{c|}{SDXS}                                           & \multicolumn{3}{c|}{PixArt}                             & \multicolumn{3}{c|}{SDv1.5}                                            & \multicolumn{3}{c|}{SDXL}                               & \multicolumn{3}{c}{Average}                                                              \\
\multicolumn{1}{l}{}              &                          & auroc $\uparrow$             & aupr $\uparrow$              & fpr95$\downarrow$              & auroc $\uparrow$         & aupr $\uparrow$          & fpr95$\downarrow$          & auroc $\uparrow$         & aupr $\uparrow$          & fpr95$\downarrow$                         & auroc $\uparrow$         & aupr $\uparrow$          & fpr95$\downarrow$          & \multicolumn{1}{c}{auroc $\uparrow$} & \multicolumn{1}{c}{aupr $\uparrow$} & \multicolumn{1}{c}{fpr95$\downarrow$} \\
\midrule

                                  & \method~ ROUGE (LLAVA) (1\_recall)  & 69.23\%              & 67.31\%              & 71.63\%              & 90.36\%          & 92.19\%          & 51.00\%          & 77.52\%          & 75.82\%          & 64.28\%                         & 64.80\%          & 58.63\%          & 67.02\%          & 75.48\%                      & 73.49\%                     & 63.48\%                     \\
                                  & \method~ ROUGE (LLAVA) (L\_recall)  & 60.80\%              & 61.53\%              & 79.12\%              & 88.80\%          & 89.13\%          & 46.95\%          & 67.08\%          & 68.11\%          & 75.19\%                         & 59.14\%          & 55.06\%          & 73.28\%          & 68.95\%                      & 68.46\%                     & 68.64\%                     \\
                                  & \method~BERT (LLAVA) (recall)      & 69.01\%              & 71.06\%              & 77.56\%              & 92.94\%          & 93.74\%          & 32.58\%          & 74.65\%          & 76.73\%          & 74.55\%                         & 61.53\%          & 58.76\%          & 79.43\%          & 74.53\%                      & 75.07\%                     & 66.03\%                     \\
\cmidrule{2-17}
                                 & \method~ ROUGE (llama) (1\_recall)  & 30.06\%              & 49.70\%              & 99.82\%              & 58.90\%          & 69.34\%          & 94.50\%          & 43.24\%          & 63.57\%          & 99.94\%                         & 52.46\%          & 64.84\%          & 99.14\%          & 46.17\%                      & 61.86\%                     & 98.35\%                     \\
                                & \method~ ROUGE (llama) (L\_recall)  & 29.78\%              & 49.81\%              & 99.69\%              & 58.95\%          & 68.82\%          & 94.00\%          & 46.79\%          & 66.07\%          & 99.59\%                         & 52.51\%          & 65.62\%          & 98.99\%          & 47.01\%                      & 62.58\%                     & 98.07\%                     \\
                                & \method~BERT (llama) (recall)      & 56.27\%              & 60.44\%              & 87.53\%              & 66.82\%          & 72.59\%          & 83.09\%          & 65.91\%          & 76.15\%          & 82.22\%                         & 62.67\%          & 69.48\%          & 86.46\%          & 62.92\%                      & 69.66\%                     & 84.82\%                     \\
\cmidrule{2-17}
                                  & \method~ ROUGE (Molmo)(1\_recall)  & 84.00\%              & 85.10\%              & 57.07\%              & 95.47\%          & 97.06\%          & 30.84\%          & 87.24\%          & 90.66\%          & 48.90\%                         & 83.56\%          & 84.31\%          & 52.78\%          & 87.57\%                      & 89.28\%                     & 47.40\%                     \\
                                  & \method~ ROUGE (Molmo)(L\_recall)  & 87.64\%              & 89.56\%              & 55.28\%              & 96.96\%          & 98.02\%          & 16.85\%          & \textbf{90.20\%} & \textbf{93.40\%} & \textbf{43.56\%}                & \textbf{88.19\%} & \textbf{89.84\%} & \textbf{47.44\%} & \textbf{90.75\%}             & \textbf{92.71\%}            & \textbf{40.78\%}            \\

\multirow{-10}{*}{\rotatebox{90}{microscopic}}    & \method~ BERT (Molmo)(recall)      & \textbf{88.45\%}     & \textbf{89.71\%}     & \textbf{48.68\%}     & \textbf{97.83\%} & \textbf{98.41\%} & \textbf{10.32\%} & 87.75\%          & 91.70\%          & 51.03\%                         & 83.10\%          & 85.46\%          & 58.14\%          & 89.28\%                      & 91.32\%                     & 42.04\%                     \\
\midrule 

                                  & \method~ ROUGE (LLAVA) (1\_recall)  & 58.61\%              & 58.86\%              & 93.11\%              & 92.32\%          & 94.24\%          & 52.38\%          & 50.93\%          & 52.14\%          & 94.84\%                         & 44.51\%          & 48.53\%          & 97.63\%          & 61.59\%                      & 63.44\%                     & 84.49\%                     \\
                                 & \method~ ROUGE (LLAVA) (L\_recall)  & 58.89\%              & 61.82\%              & 97.65\%              & 94.17\%          & 95.49\%          & 35.64\%          & 53.26\%          & 54.81\%          & 96.51\%                         & 54.03\%          & 54.40\%          & 95.88\%          & 65.09\%                      & 66.63\%                     & 81.42\%                     \\
                                & \method~ BERT (LLAVA)(recall)      & 48.89\%              & 53.08\%              & 95.31\%              & 89.16\%          & 90.98\%          & 57.78\%          & 41.42\%          & 47.63\%          & 97.96\% & 33.93\%          & 42.77\%          & 98.98\%          & 53.35\%                      & 58.62\%                     & 87.51\%                     \\
\cmidrule{2-17}                                
                                & \method~ ROUGE (llama) (1\_recall)  & 34.47\%              & 49.21\%              & 99.93\%              & 70.89\%          & 73.16\%          & 84.22\%          & 51.95\%          & 59.10\%          & 99.95\%                         & 51.17\%          & 60.32\%          & 99.54\%          & 52.12\%                      & 60.45\%                     & 95.91\%                     \\
                                & \method~ ROUGE (llama) (L\_recall)  & 35.45\%              & 50.23\%              & 99.83\%              & 72.00\%          & 74.78\%          & 80.15\%          & 55.99\%          & 61.70\%          & 99.44\%                         & 51.60\%          & 62.11\%          & 99.45\%          & 53.76\%                      & 62.21\%                     & 94.72\%                     \\
                                & \method~BERT (llama) (recall)      & 51.96\%              & 57.38\%              & 95.65\%              & 70.80\%          & 73.65\%          & 92.46\%          & 60.68\%          & 60.96\%          & 91.74\%                         & 51.64\%          & 56.84\%          & 93.57\%          & 58.77\%                      & 62.20\%                     & 93.35\%                     \\
                                \cmidrule{2-17}
                                  & \method~ ROUGE (Molmo)(1\_recall)  & 85.59\%              & 86.59\%              & 63.87\%              & 94.14\%          & 95.97\%          & 49.29\%          & 73.24\%          & 72.93\%          & 79.56\%                         & 73.07\%          & 74.54\%          & 83.48\%          & 81.51\%                      & 82.51\%                     & 69.05\%                     \\
                                  & \method~ ROUGE (Molmo)(L\_recall)  & \textbf{89.85\%}     & \textbf{90.63\%}     & \textbf{51.96\%}     & \textbf{97.01\%} & \textbf{97.94\%} & \textbf{16.08\%} & 83.63\%          & 84.65\%          & \textbf{65.35\%}                & \textbf{85.68\%} & \textbf{87.09\%} & \textbf{64.00\%} & \textbf{89.04\%}             & \textbf{90.08\%}            & \textbf{49.34\%}            \\
                 
\multirow{-10}{*}{\rotatebox{90}{remote sensing}} & \method~ BERT (Molmo)(recall)      & 80.50\%              & 82.56\%              & 75.19\%              & 91.93\%          & 93.08\%          & 43.30\%          & 61.27\%          & 64.74\%          & 90.08\%                         & 60.71\%          & 65.32\%          & 90.94\%          & 73.60\%                      & 76.42\%                     & 74.88\%                     \\
\midrule      
& \method~ ROUGE (LLAVA) (1\_recall)  & 27.39\%              & 66.97\%              & 99.68\%              & 82.94\%          & 94.90\%          & 84.90\%          & 31.87\%          & 68.29\%          & 91.28\%                         & 28.93\%          & 67.95\%          & 99.15\%          & 42.78\%                      & 74.53\%                     & 93.75\%                     \\
& \method~ ROUGE (LLAVA) (L\_recall)  & 28.95\%              & 67.52\%              & 99.36\%              & 76.61\%          & 92.26\%          & 87.72\%          & 31.42\%          & 67.96\%          & 91.40\%                         & 27.29\%          & 66.86\%          & 99.27\%          & 41.07\%                      & 73.65\%                     & 94.44\%                     \\
& \method~BERT (LLAVA) (recall)      & 34.28\%              & 71.87\%              & 99.15\%              & 82.56\%          & 94.51\%          & 75.89\%          & 37.21\%          & 72.56\%          & 92.46\%                         & 32.34\%          & 70.87\%          & 99.29\%          & 46.60\%                      & 77.45\%                     & 91.70\%                     \\
\cmidrule{2-17}
& \method~ ROUGE (llama) (1\_recall)  & 42.02\%              & 75.94\%              & 99.22\%              & 65.48\%          & 87.24\%          & 89.52\%          & 47.66\%          & 80.02\%          & 99.96\%                         & 51.95\%          & 80.22\%          & 97.35\%          & 51.78\%                      & 80.86\%                     & 96.51\%                     \\
& \method~ ROUGE (llama) (L\_recall)  & 41.84\%              & 76.44\%              & 98.90\%              & 65.09\%          & 87.38\%          & 91.05\%          & 50.45\%          & 81.30\%          & 99.70\%                         & 52.43\%          & 81.48\%          & 98.24\%          & 52.45\%                      & 81.65\%                     & 96.97\%                     \\
& \method~BERT (llama) (recall)      & \textbf{66.38\%}     & \textbf{85.27\%}     & \textbf{80.35\%}     & 75.62\%          & 90.81\%          & 77.67\%          & 70.99\%          & \textbf{87.77\%} & \textbf{76.84\%}                & 61.85\%          & \textbf{83.97\%} & \textbf{81.27\%} & 68.71\%                      & \textbf{86.95\%}            & \textbf{79.03\%}            \\
\cmidrule{2-17}
    & \method~ ROUGE (Molmo)(1\_recall)  & 40.77\%              & 72.88\%              & 95.54\%              & 83.26\%          & 94.80\%          & 73.48\%          & 46.61\%          & 75.64\%          & \textbf{90.87}\%                         & 49.16\%          & 78.04\%          & 93.57\%          & 54.95\%                      & 80.34\%                     & \textbf{88.36\%}                     \\
    & \method~ ROUGE (Molmo)(L\_recall)  & 45.64\%              & 76.90\%              & 95.99\%              & 84.10\%          & 95.13\%          & 74.21\%          & 47.60\%          & 77.70\%          & 93.14\%                         & 50.71\%          & \textbf{80.05\%}          & 94.35\%          & 57.01\%                      & 82.44\%                     & 89.42\%                     \\
\multirow{-10}{*}{\rotatebox{90}{Texture}}        & \method~ BERT (Molmo)(recall)      & 49.11\%              & \textbf{79.11\%}              & \textbf{94.65\%}              & \textbf{84.40\%} & \textbf{95.14\%} & \textbf{69.99\%} & 47.60\%          & \textbf{78.80\%}          & 93.85\%                         & 46.72\%          & 79.20\%          & 96.87\%          & 56.96\%                      & \textbf{83.06}\%                     & 88.84\%\\
    \bottomrule
\end{tabular}
    }
    \caption{\textbf{Performance comparison of \method~ applied to different Out of Distribution concepts and different text-to-image models}}
    \label{tab:Epistemic_supp}
\end{table*}

\begin{table*}[tbhp!]
\centering
    \resizebox{\textwidth}{!}
    {
\begin{tabular}{cl|ccc|ccc|ccc|ccc|ccc}
\toprule
\multicolumn{2}{l}{}                                         & \multicolumn{3}{c|}{SDXS}                                           & \multicolumn{3}{c|}{PixArt}                             & \multicolumn{3}{c|}{SDv1.5}                                            & \multicolumn{3}{c|}{SDXL}                               & \multicolumn{3}{c}{Average}                                                              \\
\multicolumn{1}{l}{}              &                          & auroc $\uparrow$             & aupr $\uparrow$              & fpr95$\downarrow$              & auroc $\uparrow$         & aupr $\uparrow$          & fpr95$\downarrow$          & auroc $\uparrow$         & aupr $\uparrow$          & fpr95$\downarrow$                         & auroc $\uparrow$         & aupr $\uparrow$          & fpr95$\downarrow$          & \multicolumn{1}{c}{auroc $\uparrow$} & \multicolumn{1}{c}{aupr $\uparrow$} & \multicolumn{1}{c}{fpr95$\downarrow$} \\
\midrule

\multirow{9}{*}{\rotatebox{90}{Vague}} & \method~ ROUGE (LLAVA) (1\_precision) & \textbf{99.99\%} & \textbf{100.00\%} & 0.00\% & \textbf{100.00\%} & \textbf{100.00\%} & 0.00\% & \textbf{100.00\%} & \textbf{100.00\%} & 0.00\% & \textbf{99.99\%} & \textbf{100.00\%} & 0.00\% & \multicolumn{1}{r}{\textbf{100.00\%}} & \multicolumn{1}{r}{\textbf{100.00\%}} & \multicolumn{1}{r}{0.00\%} \\
  & \method~ ROUGE (LLAVA) (L\_precision) & \textbf{99.99\%} & \textbf{100.00\%} & 0.00\% & \textbf{100.00\%} & \textbf{100.00\%} & 0.00\% & \textbf{100.00\%} & \textbf{100.00\%} & 0.00\% & \textbf{99.99\%} & \textbf{100.00\%} & 0.00\% & \multicolumn{1}{r}{\textbf{100.00\%}} & \multicolumn{1}{r}{\textbf{100.00\%}} & \multicolumn{1}{r}{0.00\%} \\
  & \method~ BERT (LLAVA) (precision) & 99.54\% & 99.95\% & 1.70\% & 99.79\% & 99.98\% & 0.30\% & 97.78\% & 99.75\% & 9.70\% & 98.92\% & 99.92\% & 5.60\% & \multicolumn{1}{r}{99.01\%} & \multicolumn{1}{r}{99.90\%} & \multicolumn{1}{r}{4.33\%} \\
  \cmidrule{2-17}
  & \method~ ROUGE (llama)(1\_precision) & 97.61\% & 99.58\% & 12.15\% & 99.31\% & 99.89\% & 2.05\% & 99.17\% & 99.85\% & 3.05\% & 98.68\% & 99.81\% & 4.50\% & \multicolumn{1}{r}{98.69\%} & \multicolumn{1}{r}{99.78\%} & \multicolumn{1}{r}{5.44\%} \\
  & \method~ ROUGE (llama)(L\_precision) & 96.03\% & 99.35\% & 19.50\% & 98.55\% & 99.77\% & 6.05\% & 96.96\% & 99.51\% & 13.45\% & 97.80\% & 99.67\% & 10.20\% & \multicolumn{1}{r}{97.34\%} & \multicolumn{1}{r}{99.58\%} & \multicolumn{1}{r}{12.30\%} \\
  & \method~ BERT (llama)(precision) & 60.63\% & 92.24\% & \textbf{82.85\%} & 67.02\% & 94.54\% & 83.10\% & 37.35\% & 86.51\% & 92.70\% & 70.85\% & 95.13\% & 78.00\% & \multicolumn{1}{r}{58.96\%} & \multicolumn{1}{r}{92.11\%} & \multicolumn{1}{r}{\textbf{84.16\%}} \\
  \cmidrule{2-17}
 & \method~ ROUGE (Molmo)(1\_precision) & \textbf{99.99\%} & \textbf{100.00\%} & 0.00\% & \textbf{100.00\%} & \textbf{100.00\%} & \textbf{0.00\%} & \textbf{99.99\%} & \textbf{100.00\%} & \textbf{0.00\%} & \textbf{99.99\%} & \textbf{100.00\%} & 0.00\% & \multicolumn{1}{r}{\textbf{99.99\%}} & \multicolumn{1}{r}{\textbf{100.00\%}} & \multicolumn{1}{r}{0.00\%} \\
 & \method~ ROUGE (Molmo)(L\_precision) & \textbf{99.99\%} & \textbf{100.00\%} & 0.00\% & \textbf{100.00\%} & \textbf{100.00\%} & \textbf{0.00\%} & \textbf{99.99\%} & \textbf{100.00\%} & \textbf{0.00\%} & \textbf{100.00\%} & \textbf{100.00\%} & 0.00\% & \multicolumn{1}{r}{\textbf{100.00\%}} & \multicolumn{1}{r}{\textbf{100.00\%}} & \multicolumn{1}{r}{0.00\%} \\
 & \method~  BERT (Molmo)(precision) & 60.63\% & 92.24\% & 82.85\% & 67.02\% & 94.54\% & 83.10\% & 37.35\% & 86.51\% & 92.70\% & 70.85\% & 95.13\% & 78.00\% & \multicolumn{1}{r}{58.96\%} & \multicolumn{1}{r}{92.11\%} & \multicolumn{1}{r}{84.16\%} \\
 \midrule 
\multirow{9}{*}{\rotatebox{90}{Adversarial}} & \method~ ROUGE (LLAVA) (1\_precision) & \textit{70.83\%} & \textit{98.42\%} & \textit{47.80\%} & \textit{75.17\%} & \textit{98.83\%} & \textit{48.26\%} & \textit{59.03\%} & \textit{96.59\%} & \textit{49.60\%} & \textit{65.03\%} & \textit{98.31\%} & \textit{43.99\%} & 67.52\% & 98.04\% & 47.41\% \\
 & \method~ ROUGE (LLAVA) (L\_precision) & \textit{71.61\%} & \textit{98.56\%} & \textit{49.00\%} & \textit{74.33\%} & \textit{98.75\%} & \textit{48.26\%} & 59.49\% & 96.80\% & 50.00\% & 64.87\% & 98.38\% & \textbf{43.99\%} & 67.58\% & 98.12\% & 47.81\% \\
 & \method~ BERT (LLAVA)(precision) & \textit{\textbf{79.14\%}} & \textit{\textbf{99.20\%}} & \textit{\textbf{48.00\%}} & \textit{74.70\%} & \textit{98.84\%} & \textit{48.06\%} & \textit{61.21\%} & \textit{97.41\%} & \textit{54.20\%} & \textit{66.73\%} & \textit{98.78\%} & \textit{48.11\%} & \textbf{70.44\%} & \textbf{98.56\%} & \textbf{49.59\%} \\
\cmidrule{2-17}
 & \method~ ROUGE (llama)(1\_precision) & \textit{64.83\%} & \textit{98.18\%} & \textit{57.03\%} & \textit{78.08\%} & \textit{99.11\%} & \textit{49.20\%} & \textit{67.27\%} & \textit{98.20\%} & \textit{51.50\%} & \textit{\textbf{69.57\%}} & \textit{\textbf{98.40\%}} & \textit{50.60\%} & 69.94\% & 98.47\% & 52.08\% \\
 & \method~ ROUGE (llama)(L\_precision) & \textit{69.71\%} & \textit{98.49\%} & \textit{59.64\%} & \textit{\textbf{80.99\%}} & \textit{\textbf{99.23\%}} & \textit{\textbf{49.00\%}} & \textit{\textbf{68.79\%}} & \textit{\textbf{98.33\%}} & \textit{56.11\%} & \textit{71.14\%} & \textit{98.47\%} & \textit{53.40\%} & \textbf{72.66\%} & \textbf{98.63\%} & \textbf{54.54\%} \\
 & \method~ BERT (llama)(precision) & \textit{73.98\%} & \textit{98.72\%} & \textit{78.31\%} & \textit{72.95\%} & \textit{98.93\%} & \textit{86.40\%} & \textit{45.09\%} & \textit{96.93\%} & \textit{94.18\%} & \textit{72.69\%} & \textit{98.80\%} & \textit{81.40\%} & 66.18\% & 98.34\% & 85.07\% \\
\cmidrule{2-17}
 & \method~ ROUGE (Molmo)(1\_precision) & \textit{64.14\%} & \textit{97.77\%} & \textit{48.80\%} & \textit{76.32\%} & \textit{98.90\%} & \textit{48.00\%} & \textit{63.59\%} & \textit{97.73\%} & \textit{49.50\%} & \textit{62.97\%} & \textit{97.64\%} & \textit{49.20\%} & 66.76\% & 98.01\% & 48.87\% \\
 & \method~ ROUGE (Molmo)(L\_precision) & \textit{63.20\%} & \textit{97.75\%} & \textit{49.00\%} & \textit{76.26\%} & \textit{98.90\%} & \textit{48.40\%} & \textit{64.09\%} & \textit{97.90\%} & \textit{\textbf{49.70\%}} & \textit{63.76\%} & \textit{97.87\%} & \textit{49.60\%} & 66.83\% & 98.11\% & 49.17\% \\
 & \method~  BERT (Molmo)(precision) & \textit{73.98\%} & \textit{98.72\%} & \textit{78.31\%} & \textit{72.95\%} & \textit{98.93\%} & \textit{86.40\%} & \textit{45.09\%} & \textit{96.93\%} & \textit{94.18\%} & \textit{72.69\%} & \textit{98.80\%} & \textit{81.40\%} & 66.18\% & 98.34\% & 85.07\% \\
 \midrule
\multirow{9}{*}{\rotatebox{90}{Corrupt lvl1}} & \method~ ROUGE (LLAVA) (1\_precision) & 44.38\% & 46.02\% & 96.50\% & 51.31\% & 53.87\% & 96.51\% & 19.90\% & 34.21\% & 99.22\% & 19.60\% & 47.23\% & 99.52\% & 33.80\% & 45.33\% & 97.94\% \\
  & \method~ ROUGE (LLAVA) (L\_precision) & 46.06\% & 48.93\% & 97.12\% & 48.58\% & 51.26\% & 96.61\% & 18.98\% & 34.00\% & 99.49\% & 18.26\% & 46.95\% & 99.72\% & 32.97\% & 45.29\% & 98.24\% \\
  & \method~ BERT (LLAVA)(precision) & 60.71\% & 65.46\% & 94.87\% & 49.73\% & 52.23\% & 96.63\% & 25.07\% & 36.22\% & 99.62\% & 25.65\% & 50.22\% & 99.57\% & 40.29\% & 51.03\% & 97.67\% \\
\cmidrule{2-17}
  & \method~ ROUGE (llama)(1\_precision) & 33.10\% & 43.04\% & 99.88\% & 58.67\% & 62.73\% & 97.71\% & 35.83\% & 42.27\% & 99.14\% & 42.49\% & 46.81\% & 99.77\% & 42.52\% & 48.71\% & 99.13\% \\
  & \method~ ROUGE (llama)(L\_precision) & 44.92\% & 48.73\% & 98.85\% & 64.78\% & 65.93\% & 92.88\% & 39.64\% & 43.63\% & 98.08\% & 47.27\% & 48.85\% & 98.79\% & 49.15\% & 51.79\% & 97.15\% \\
  & \method~ BERT (llama)(precision) & \textbf{85.85\%} & \textbf{84.45\%} & \textbf{73.94\%} & \textbf{79.43\%} & \textbf{78.83\%} & \textbf{87.21\%} & \textbf{54.97\%} & \textbf{53.73\%} & \textbf{94.51\%} & \textbf{75.76\%} & \textbf{72.91\%} & \textbf{84.65\%} & \textbf{74.00\%} & \textbf{72.48\%} & \textbf{85.08\%} \\
\cmidrule{2-17}
 & \method~ ROUGE (Molmo)(1\_precision) & 28.75\% & 38.34\% & 98.77\% & 54.58\% & 55.29\% & 96.02\% & 27.69\% & 37.57\% & 98.78\% & 27.45\% & 37.27\% & 98.54\% & 34.62\% & 42.12\% & 98.03\% \\
 & \method~ ROUGE (Molmo)(L\_precision) & 27.29\% & 38.11\% & 99.15\% & 54.32\% & 55.59\% & 96.16\% & 27.30\% & 38.07\% & 99.45\% & 27.35\% & 37.89\% & 99.37\% & 34.07\% & 42.42\% & 98.53\% \\
 & \method~  BERT (Molmo)(precision) & \textbf{85.85\%} & \textbf{84.45\%} & \textbf{73.94\%} & \textbf{79.43\%} & \textbf{78.83\%} & 87.21\% & \textbf{54.97\%} & \textbf{53.73\%} & 94.51\% & \textbf{75.76\%} & \textbf{72.91\%} & 84.65\% & \textbf{74.00\%} & \textbf{72.48\%} & \textbf{85.08\%} \\
 \midrule 
\multirow{9}{*}{\rotatebox{90}{Corrupt lvl2}}  & \method~ ROUGE (LLAVA) (1\_precision) & \textit{25.37\%} & \textit{35.74\%} & \textit{98.11\%} & \textit{62.14\%} & \textit{62.43\%} & \textit{90.66\%} & \textit{26.18\%} & \textit{36.05\%} & \textit{98.13\%} & \textit{27.77\%} & \textit{50.14\%} & \textit{97.85\%} & 35.37\% & 46.09\% & 96.19\% \\
  & \method~ ROUGE (LLAVA) (L\_precision) & \textit{25.47\%} & \textit{35.87\%} & \textit{98.41\%} & \textit{60.56\%} & \textit{60.22\%} & \textit{90.25\%} & \textit{25.45\%} & \textit{35.94\%} & \textit{98.69\%} & \textit{26.30\%} & \textit{49.90\%} & \textit{98.56\%} & 34.45\% & 45.48\% & 96.48\% \\
  & \method~ BERT (LLAVA)(precision) & \textit{33.19\%} & \textit{39.35\%} & \textit{98.28\%} & \textit{66.19\%} & \textit{66.09\%} & \textit{86.84\%} & \textit{33.18\%} & \textit{39.55\%} & \textit{98.58\%} & \textit{37.58\%} & \textit{55.84\%} & \textit{97.01\%} & 42.54\% & 50.21\% & 95.18\% \\
\cmidrule{2-17}
  & \method~ ROUGE (llama)(1\_precision) & \textit{33.29\%} & \textit{42.84\%} & \textit{99.90\%} & \textit{58.79\%} & \textit{62.71\%} & \textit{97.92\%} & \textit{35.71\%} & \textit{42.10\%} & \textit{99.14\%} & \textit{42.49\%} & \textit{46.84\%} & \textit{99.78\%} & 42.57\% & 48.62\% & 99.19\% \\
  & \method~ ROUGE (llama)(L\_precision) & \textit{45.22\%} & \textit{48.60\%} & \textit{98.82\%} & \textit{64.79\%} & \textit{65.81\%} & \textit{92.98\%} & \textit{39.54\%} & \textit{43.43\%} & \textit{98.18\%} & \textit{47.04\%} & \textit{48.77\%} & \textit{98.76\%} & 49.15\% & 51.65\% & 97.19\% \\
  & \method~ BERT (llama)(precision) & \textit{\textbf{85.96\%}} & \textit{\textbf{84.23\%}} & \textit{\textbf{73.54\%}} & \textit{\textbf{79.47\%}} & \textit{\textbf{78.80\%}} & \textit{\textbf{87.05\%}} & \textit{54.78\%} & \textit{53.51\%} & \textit{94.79\%} & \textit{\textbf{76.04\%}} & \textit{\textbf{73.44\%}} & \textit{\textbf{84.58\%}} & \textbf{74.06\%} & \textbf{72.49\%} & \textbf{84.99\%} \\
\cmidrule{2-17}
 & \method~ ROUGE (Molmo)(1\_precision) & \textit{28.76\%} & \textit{38.10\%} & \textit{98.59\%} & \textit{54.77\%} & \textit{55.49\%} & \textit{96.02\%} & \textit{27.66\%} & \textit{37.42\%} & \textit{98.75\%} & \textit{27.39\%} & \textit{37.23\%} & \textit{98.53\%} & 34.64\% & 42.06\% & 97.97\% \\
 & \method~ ROUGE (Molmo)(L\_precision) & \textit{27.33\%} & \textit{37.87\%} & \textit{99.16\%} & \textit{54.51\%} & \textit{55.87\%} & \textit{96.11\%} & \textit{27.31\%} & \textit{37.92\%} & \textit{99.35\%} & \textit{27.33\%} & \textit{37.83\%} & \textit{99.36\%} & 34.12\% & 42.37\% & 98.50\% \\
 & \method~  BERT (Molmo)(precision) & \textit{\textbf{85.96\%}} & \textit{\textbf{84.23\%}} & \textit{\textbf{73.54\%}} & \textit{\textbf{79.47\%}} & \textit{\textbf{78.80\%}} & \textit{\textbf{87.05\%}} & \textit{54.78\%} & \textit{53.51\%} & \textit{94.79\%} & \textit{\textbf{76.04\%}} & \textit{\textbf{73.44\%}} & \textit{\textbf{84.58\%}} & \textbf{74.06\%} & \textbf{72.49\%} & \textbf{84.99\%} \\ 
     \bottomrule

\end{tabular}
}
    \caption{
        \textbf{Performance comparison all applications (\textit{Vague}, \textit{Adversarial}, \textit{Corrupt Lvl1},\textit{ Corrup Lvl2})} \label{tab:Aleatoric_supp}}
\end{table*}

\section{Dataset of prompts}\label{sec:Dataset_Supp}
 We need an In Distribution dataset of prompt and  Distribution datasets of prompt with uncertainty. For the In Distribution dataset of prompt we have choose to use the dataset proposed by \cite{ding2023gpt4image}, where the authors generate high-quality descriptions for each training image of ImageNet \cite{deng2009imagenet} by interacting with GPT-4. These descriptions include detailed prompt, providing a high information about the image content. We have choosen to use prompt of ImageNet images because this images are use as pretrained of most diffusion model since a lot of them use DiT \cite{peebles2023scalable} backbone or a variant of this diffusion model. We denote this dataset of prompt as Normal.

Regarding the Uncertainty prompt, since we have two kind of uncertainty we have mainly to kind to build the prompt.
 First regarding the OOD dataset of prompt we have used images of remote sensing~\cite{BigEarthNet}, texture~\cite{cimpoi14describing}, Microscopic~\cite{Microdata}  datasets. The idea is that diffusion model trained on LAION-5b dataset \cite{schuhmann2022laion} often lack good perfomences on this dataset due to the fact that this kind of images are quite absent on  LAION-5b. Then once the we have collected the images, we used LLAVA next for its efficiency and velocity to caption the images. We denote these dataset of prompt  respectively as Remote sensing, Texture, and Microscopic.

 Regarding the aleatoric uncertainty prompt, we proposes two datasets. One called vague which is composed of 2k prompts of the following shape: "An image of ***."  "An picture of ***." where we replaced the *** with the name of the class of ImageNet. We denote this dataset of prompt as Vague.


To conduct our experiments, we require both an \textit{in-distribution dataset of prompts }and d\textit{atasets of prompts with uncertainty}.

\paragraph{In-Distribution Dataset of Prompts}
 For our in-distribution dataset, we use the dataset proposed by \cite{ding2023gpt4image}, in which the authors generate high-quality descriptions for each training image in the ImageNet dataset \cite{deng2009imagenet} by interacting with GPT-4. These descriptions provide detailed and contextually rich prompts that thoroughly represent the image content. We selected randomly 20 by class prompts derived from ImageNet images because many diffusion models are pretrained on ImageNet, given that they often utilize a DiT backbone or variants of this architecture. Using prompts that align with the content on which these models are pretrained ensures the consistency of the in-distribution data. We refer to this dataset as \textit{Normal}.

\paragraph{Out-of-Distribution (OOD) Prompts for Epistemic Uncertainty}
For prompts representing epistemic uncertainty, we need out-of-distribution (OOD) content, as this type of uncertainty arises when models encounter unfamiliar or untrained data. For this purpose, we selected images from domains typically underrepresented in the LAION dataset \cite{schuhmann2022laion}, which many diffusion models are trained on. These domains include remote sensing, texture, and microscopic images, which are relatively rare in the Laion-5b dataset \cite{schuhmann2022laion}, potentially leading to poorer model performance on these types. After collecting images from the test sets of remote sensing \cite{BigEarthNet}, texture \cite{cimpoi14describing}, and microscopic datasets \cite{Microdata}, we used the LLAVA Next model \cite{li2024llava} to caption each image. LLAVA Next was chosen for its efficiency and speed in generating relevant descriptions for images from diverse, specialized domains. We label these OOD prompt datasets as \textit{Remote Sensing}, \textit{Texture}, and \textit{Microscopic}.

\paragraph{Aleatoric Uncertainty Prompts}
 For prompts that simulate aleatoric uncertainty, which stems from inherent noise or ambiguity in the prompt, we designed two datasets:
 \begin{enumerate}
    \item  \textit{Vague}: This dataset contains 2,000 prompts with deliberately vague descriptions, structured to provide minimal context, such as "An image of ***" or "A picture of ***", where "***" is replaced by the ImageNet class name. These prompts create an ambiguous context, simulating scenarios where the input information is too sparse for the model to fully comprehend. We refer to this dataset as \textit{Vague}.
    \item \textit{Adversarial}: This dataset contains 1,000 prompts altered from the \textit{Normal} dataset using UnlearnDiffAtk \cite{zhang2024generatenotsafetydrivenunlearned}, a gradient-based adversarial attack method optimizing adversarial prompts within the diffusion process.
     \item \textit{Corrupted}: To simulate real-world scenarios with input noise, we created prompts with grammatical errors and word omissions. Using LLAMA-3-2, we generated captions with varying levels of corruption:
   (1) Level 1: We introduced grammatical mistakes and spelling errors to the prompts from the \textit{Normal} dataset.
   (2) Level 2: We further intensified corruption by randomly removing 50\% of the words in each Level 1 prompt, simulating extreme cases of incomplete or fragmented input. These corrupted prompts are labeled as \textit{Corrupted} and represent varying degrees of aleatoric uncertainty that could interfere with the model’s comprehension.
\end{enumerate}

\section{Task: Knowledge extraction: Concept}
\label{sec:CopyrightPolitician}

\subsection{Deepfake Uncertainty}\label{sec:Deepfake}

An intriguing aspect of uncertainty estimation is its potential to benefit other tasks. One pressing issue with diffusion models is the challenge of protecting against harmful applications, such as deepfakes, which pose significant societal risks. As diffusion models continue to improve, the likelihood of misuse grows, raising concerns about the ability to generate realistic images of prominent individuals, such as politicians.

To investigate whether diffusion models have learned to generate images of specific politicians, we focused on the heads of state from the G7 countries. We created 100 prompts with the names of these politicians and used the four previously mentioned diffusion models to generate images based on these prompts. To quantify the uncertainty, we employed a LVLM, but with a modified prompt. Rather than asking the LVLM to describe the image, we directly queried whether a specific politician was present in the image, with a simple ``yes" or ``no" response. This approach allowed us to assess the LVLM’s ability to detect the intended concept—politician recognition—in the generated images. Our findings are summarized in Table \ref{tab:politicians}.

One key insight from this study is that the effectiveness of concept recognition depends heavily on the LVLM model’s ability to identify the targeted concept in an image. To assess Molmo's capacity for concept extraction, we evaluated it on 20 images of the selected politicians from the web, along with images of individuals who were not politicians. The results of Molmo’s performance in recognizing each politician are shown in Table~\ref{tab:politician_recognition} are as follows: 

\begin{table}[h]
\centering
\begin{tabular}{lcc}
\toprule
Name & Precision (\%) & Recall (\%) \\
\midrule
Joe Biden & 83 & 100 \\
Emmanuel Macron & 82 & 100 \\
King Charles III & 100 & 100 \\
Justin Trudeau & 88 & 100 \\
Kishida Fumio & 83 & 33 \\
Giorgia Meloni & 61 & 73 \\
Olaf Scholz & 55 & 100 \\
\bottomrule
\end{tabular}
\caption{Performance of Molmo in \textit{Politician} Recognition using Precision and Recall Measures  \label{tab:politician_recognition}}
\end{table}
These results demonstrate varying levels of effectiveness across different politicians. For example, Molmo performs particularly well in identifying King Charles III, achieving perfect precision and recall. In contrast, it struggles more with Giorgia Meloni and Olaf Scholz, where precision and recall scores are lower.

In analyzing the diffusion models, we observed that SDXL was the most consistent in generating recognizable images of Emmanuel Macron and demonstrated the highest overall accuracy in generating images for the given prompts. Visual inspection confirmed that SDXL produced the best representations for most prompts, showcasing its relative reliability in political figure generation.

\begin{table*}[tbhp!]
    \centering
    \resizebox{0.8\textwidth}{!}
    {
\begin{tabular}{lrrrrrrrr}
\toprule
        & \multicolumn{1}{l}{Justin Trudeau} & \multicolumn{1}{l}{King Charles} & \multicolumn{1}{l}{Fumio Kishida} & \multicolumn{1}{l}{Olaf Scholz} & \multicolumn{1}{l}{Emmanuel Macron} & \multicolumn{1}{l}{Joe Bidden} & \multicolumn{1}{l}{Giorgia Meloni} & \multicolumn{1}{l}{AVERAGE} \\
        \midrule 
SDXL    & 94                               & 94                             & 88                              & 88                            & 98                                & 96                           & 21                               & \textbf{95.5}              \\
SDv1.5   & 78                               & 60                              & 62                              & 88                            & 94                                & 80                            & 18                               & 78                        \\
SDXS    & 80                                & 40                              & 70                               & 82                            & 94                                & 76                           & 28                               & 72.5                       \\
PixArt  & 36                               & 76                             & 62                              & 78                            & 94                                & 90                            & 15                               & 74                        \\
AVERAGE & 72                               & 67.5                            & 70.5                             & 84                            & \textbf{95}                       & 85.5                          & 20.5                              &  70.71 \\
    \bottomrule      
\end{tabular}
    }
    \caption{
        \textbf{Performance comparison  \textit{Politicians} Application}: Accuracy (\%) Measures}
    \label{tab:politicians}
\end{table*}

\subsection{Copyright Uncertainty}\label{sec:Copyright}

Another important category of concepts worth evaluating is related to copyright. Specifically, we explore whether a diffusion model can generate well-known cartoon characters from specific brands, such as Mickey Mouse, Donald Duck, Darth Vader, and Pikachu. This experiment enables us to assess the extent to which the model has learned to generate these copyrighted concepts. 

Following a similar approach to the previous experiments, we generated 100 prompts for each character concept and used the various diffusion models to generate corresponding images. We then employed the LVLM component of \method~to determine whether the target character was present in each generated image, asking the LVLM model to provide a simple "yes" or "no" answer regarding the presence of each concept.

The results of this experiment are presented in Table \ref{tab:copyright}. Notably, while PixArt performed less effectively in the earlier experiments with politicians, it emerges as one of the top performers in this context, successfully generating recognizable representations of copyrighted characters. This contrast suggests differences in the training datasets used for these diffusion models and highlights how these variations can impact model behavior in response to regulatory constraints and copyright considerations.

\subsection{Task: Bias of Diffusion Model}\label{sec:Bias}

Previous work \cite{aldahoul2024ai}  has shown that diffusion models may exhibit gender and racial biases when generating people from vague prompts. In our experiments, we explicitly instructed diffusion models to generate individuals performing one of 13 specific jobs, including `CEO', `basketball player', `call center employee', `cleaning staff', `computer user', `firefighter', `marketing professional', `medical doctor', `nurse', `police officer', `politician', `rap singer', and `teacher'. 

We specified both the gender and race of the person to be generated, limiting the racial categories to white, Black, and Asian to simplify the analysis, as results became inconsistent with a larger variety of races. Similar to previous experiments, we used a LVLM to confirm whether the specified concepts were accurately generated using a simple yes/no question.

\begin{figure*}[t]
\centering
\includegraphics[width=0.70\linewidth]{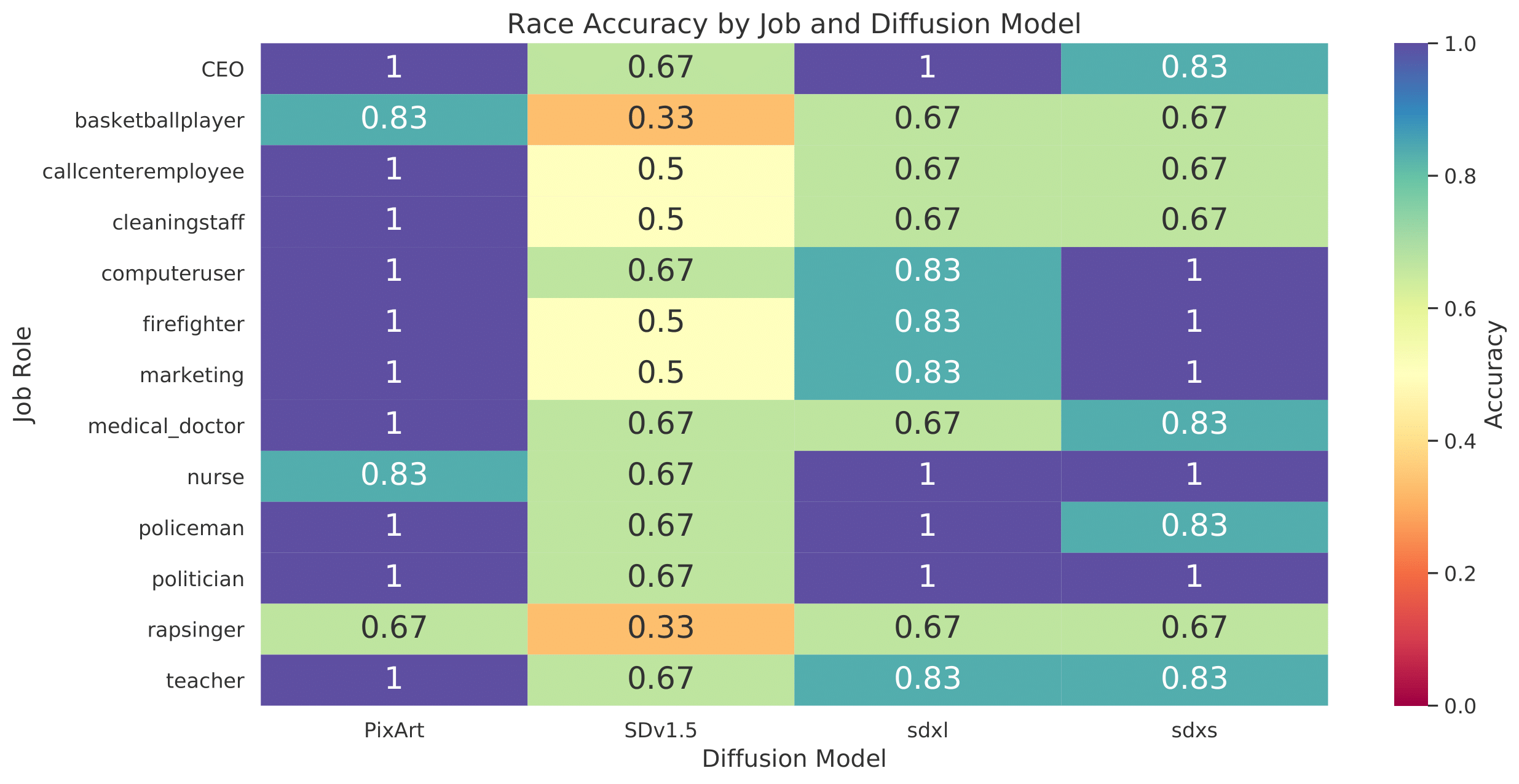}
\caption{Illustration of gender bias in diffusion models with respect to job representation.
}
\vspace{-1em}
\label{fig:biais1}
\end{figure*}

\begin{figure*}[t]
\centering
\includegraphics[width=0.70\linewidth]{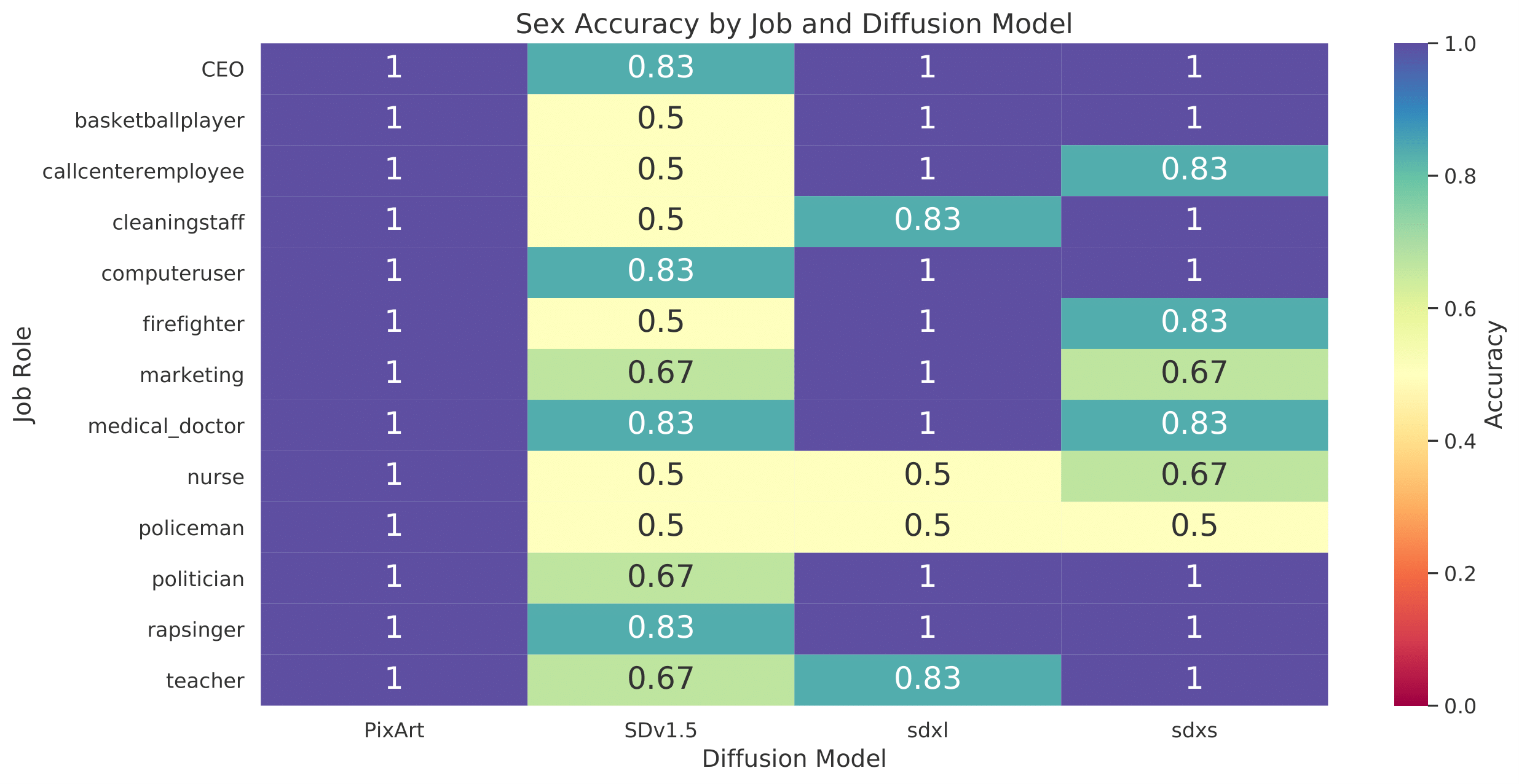}
\caption{Illustration of racial bias in diffusion models with respect to job representation.
}
\vspace{-1em}
\label{fig:biais2}
\end{figure*}

The results of this study, shown in Figures \ref{fig:biais1} and \ref{fig:biais2}, reveal significant fairness issues in SDv1.5, which tend to diminish in newer models like PixArt. Interestingly, however, PixArt still exhibits biases; for example, it struggles to generate a white basketball player or a white rap singer, while defaulting to certain racial stereotypes for these jobs. These findings suggest that understanding and addressing uncertainty in diffusion models could be an important step toward mitigating such biases.

\begin{table}[htbp!]
    \centering
    \resizebox{\columnwidth}{!}
    {
\begin{tabular}{lrrrrr}
\toprule
        & \multicolumn{1}{l}{Darth Vader} & \multicolumn{1}{l}{Pikachu} & \multicolumn{1}{l}{Mickey} & \multicolumn{1}{l}{Donald} & \multicolumn{1}{l}{Average} \\
                \midrule 
SDv1.5  & 96                            & 88                        & 86                       & 8                        & 87.5                       \\
PixArt  & 96                            & 100                           & 100                          & 94                       & 97.5                       \\
SDXS    & 100                               & 96                        & 96                       & 90                        & 95.5                       \\
SDXL    & 98                            & 100                           & 94                       & 96                       & 97                        \\
AVERAGE & 97.5                           & 96                        & 94                       & 90                        & 94.3\\
\multicolumn{1}{l}{} \\      
    \bottomrule      
\end{tabular}
}
    \caption{
        Performance of Molmo in \textit{Copyrights} Recognition: Accuracy (\%) Metrics}
    \label{tab:copyright}
\end{table}

\section{Qualitative Results}\label{sec:Qualitatifs}
\begin{figure*}[tbp!]
    \centering
    
    \caption{Example of a generated microscopic image with PixArt-$\Sigma$ and the caption generated}
    \label{fig-microscopic-pixart}
    \resizebox{0.8\textwidth}{!}{
    \begin{tabular}{p{0.2\textwidth}|p{0.15\textwidth}|p{0.65\textwidth}}
\toprule

\multirow{4}{*}{\includegraphics[width=0.2\textwidth]{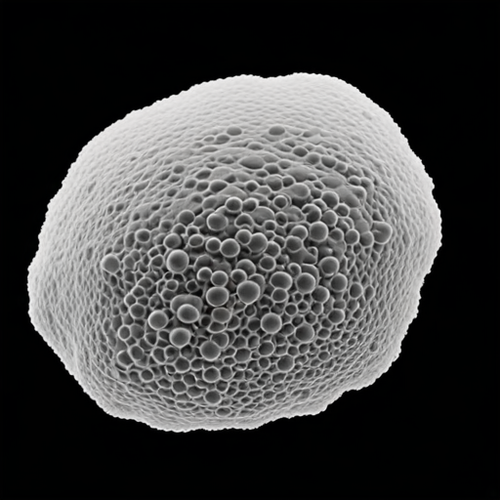}} 
& original prompt &  \small\input{images/qualitative_results/microscopic/pixart/48.txt}    \\
& llava caption   & \small\input{images/qualitative_results/microscopic/pixart/48_llava.txt}    \\
& llama caption   & \small\input{images/qualitative_results/microscopic/pixart/48_llama32.txt}    \\
& molmo caption   & \small\input{images/qualitative_results/microscopic/pixart/48_molmo.txt}    \\
\bottomrule
    \end{tabular}}
\end{figure*}

\begin{figure*}[tbph!]
    \centering
    
    \caption{Example of a generated microscopic image with SDXL and the caption generated }
    \label{fig-microscopic-sdxl}
    \resizebox{0.8\textwidth}{!}{
    \begin{tabular}{p{0.2\textwidth}|p{0.15\textwidth}|p{0.65\textwidth}}
\toprule
\multirow{4}{*}{\includegraphics[width=0.2\textwidth]{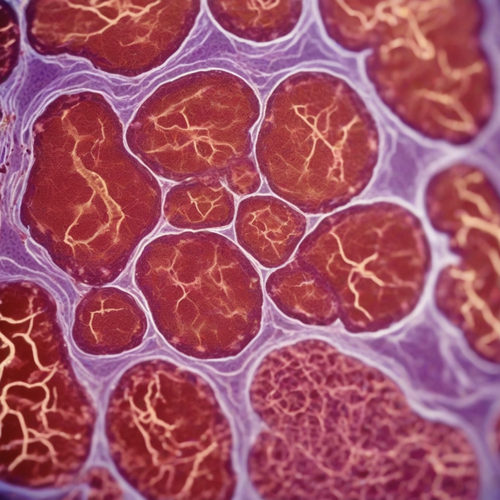}} 
& original prompt &  \small\input{images/qualitative_results/microscopic/sdxl/48.txt}    \\
& llava caption   & \small\input{images/qualitative_results/microscopic/sdxl/48_llava.txt}    \\
& llama caption   & \small\input{images/qualitative_results/microscopic/sdxl/48_llama32.txt}    \\
& molmo caption   & \small\input{images/qualitative_results/microscopic/sdxl/48_molmo.txt}    \\
\bottomrule
    \end{tabular}}
\end{figure*}

\begin{figure*}[tbp!]
    \centering
    \caption{Example of a generated microscopic image with SDXS and the caption generated }
    \label{fig-microscopic-sdxs}
    \resizebox{0.8\textwidth}{!}{
    \begin{tabular}{p{0.2\textwidth}|p{0.15\textwidth}|p{0.65\textwidth}}
\toprule
\multirow{4}{*}{\includegraphics[width=0.2\textwidth]{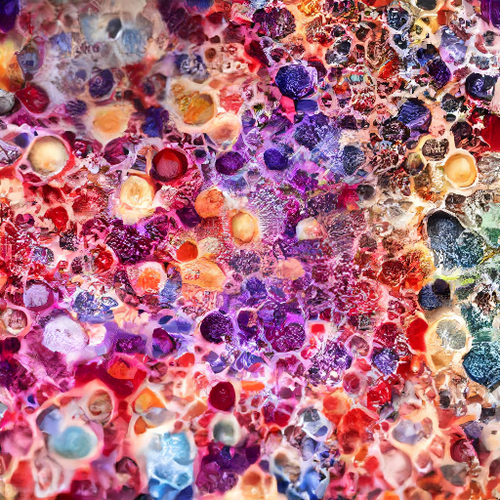}} 
& original prompt &  \small\input{images/qualitative_results/microscopic/sdxs/48.txt}    \\
& llava caption   & \small\input{images/qualitative_results/microscopic/sdxs/48_llava.txt}    \\
& llama caption   & \small\input{images/qualitative_results/microscopic/sdxs/48_llama32.txt}    \\
& molmo caption   & \small\input{images/qualitative_results/microscopic/sdxs/48_molmo.txt}    \\
\bottomrule
    \end{tabular}}
\end{figure*}

\begin{figure*}[tbp!]
    \centering
    
    \caption{Example of a generated microscopic image with SDv1.5 and the caption generated }
    \label{fig-microscopic-sdv1.5}
    \resizebox{0.8\textwidth}{!}{
    \begin{tabular}{p{0.2\textwidth}|p{0.15\textwidth}|p{0.65\textwidth}}
\toprule
\multirow{4}{*}{\includegraphics[width=0.2\textwidth]{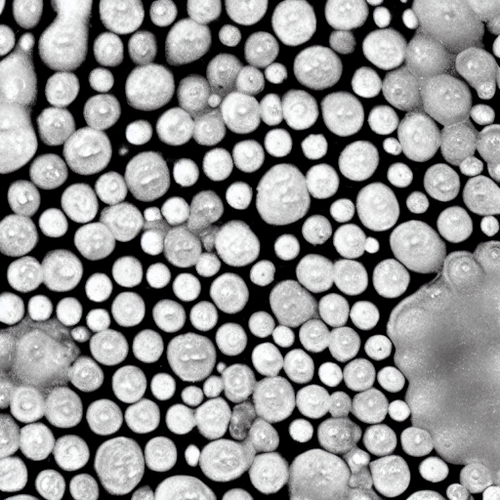}} 
& original prompt &  \small\input{images/qualitative_results/microscopic/sdv1.5/48.txt}    \\
& llava caption   & \small\input{images/qualitative_results/microscopic/sdv1.5/48_llava.txt}    \\
& llama caption   & \small\input{images/qualitative_results/microscopic/sdv1.5/48_llama32.txt}    \\
& molmo caption   & \small\input{images/qualitative_results/microscopic/sdv1.5/48_molmo.txt}    \\
\bottomrule
    \end{tabular}}
\end{figure*}

\begin{figure*}[tbp!]
    \centering
    
    \caption{Example of a generated Remote Sensing image with PixArt-$\Sigma$ and the caption generated}
    \label{fig-remote_sensing-pixart}
    \resizebox{0.8\textwidth}{!}{
    \begin{tabular}{p{0.2\textwidth}|p{0.15\textwidth}|p{0.65\textwidth}}
\toprule

\multirow{4}{*}{\includegraphics[width=0.2\textwidth]{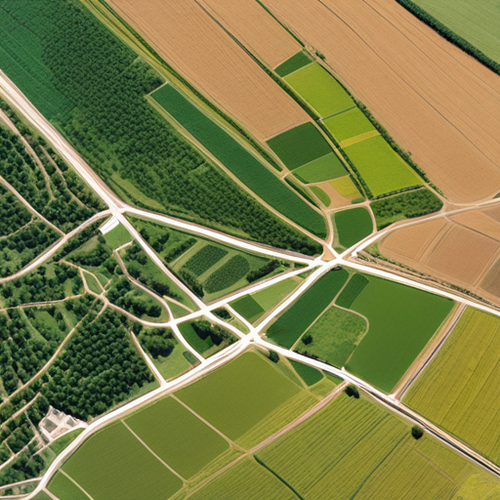}} 
& original prompt &  \small\input{images/qualitative_results/remote_sensing/pixart/12.txt}    \\
& llava caption   & \small\input{images/qualitative_results/remote_sensing/pixart/12_llava.txt}    \\
& llama caption   & \small\input{images/qualitative_results/remote_sensing/pixart/12_llama32.txt}    \\
& molmo caption   & \small\input{images/qualitative_results/remote_sensing/pixart/12_molmo.txt}    \\
\bottomrule
    \end{tabular}}
\end{figure*}

\begin{figure*}[tbph!]
    \centering
    
    \caption{Example of a generated Remote Sensing image with SDXL and the caption generated }
    \label{fig-remote_sensing-sdxl}
    \resizebox{0.8\textwidth}{!}{
    \begin{tabular}{p{0.2\textwidth}|p{0.15\textwidth}|p{0.65\textwidth}}
\toprule
\multirow{4}{*}{\includegraphics[width=0.2\textwidth]{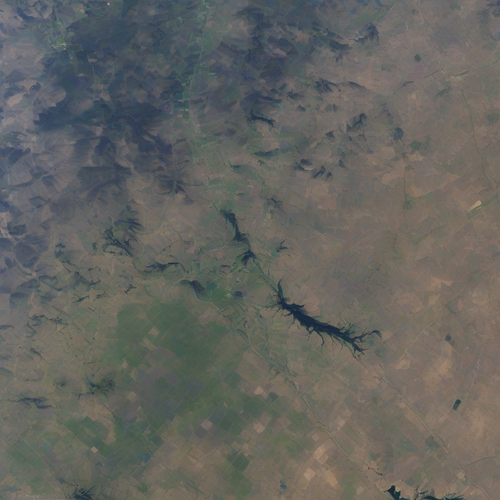}} 
& original prompt &  \small\input{images/qualitative_results/remote_sensing/sdxl/12.txt}    \\
& llava caption   & \small\input{images/qualitative_results/remote_sensing/sdxl/12_llava.txt}    \\
& llama caption   & \small\input{images/qualitative_results/remote_sensing/sdxl/12_llama32.txt}    \\
& molmo caption   & \small\input{images/qualitative_results/remote_sensing/sdxl/12_molmo.txt}    \\
\bottomrule
    \end{tabular}}
\end{figure*}

\begin{figure*}[tbp!]
    \centering
    \caption{Example of a generated Remote Sensing image with SDXS and the caption generated }
    \label{fig-remote_sensing-sdxs}
    \resizebox{0.8\textwidth}{!}{
    \begin{tabular}{p{0.2\textwidth}|p{0.15\textwidth}|p{0.65\textwidth}}
\toprule
\multirow{4}{*}{\includegraphics[width=0.2\textwidth]{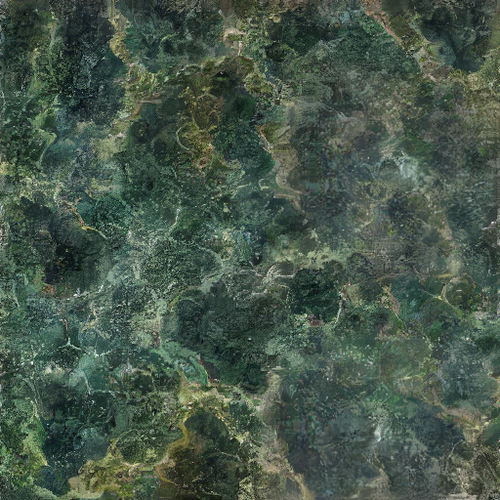}} 
& original prompt &  \small\input{images/qualitative_results/remote_sensing/sdxs/12.txt}    \\
& llava caption   & \small\input{images/qualitative_results/remote_sensing/sdxs/12_llava.txt}    \\
& llama caption   & \small\input{images/qualitative_results/remote_sensing/sdxs/12_llama32.txt}    \\
& molmo caption   & \small\input{images/qualitative_results/remote_sensing/sdxs/12_molmo.txt}    \\
\bottomrule
    \end{tabular}}
\end{figure*}

\begin{figure*}[tbp!]
    \centering
    
    \caption{Example of a generated Remote Sensing image with SD1.5 and the caption generated }
    \label{fig-remote_sensing-sdv1.5}
    \resizebox{0.8\textwidth}{!}{
    \begin{tabular}{p{0.2\textwidth}|p{0.15\textwidth}|p{0.65\textwidth}}
\toprule
\multirow{4}{*}{\includegraphics[width=0.2\textwidth]{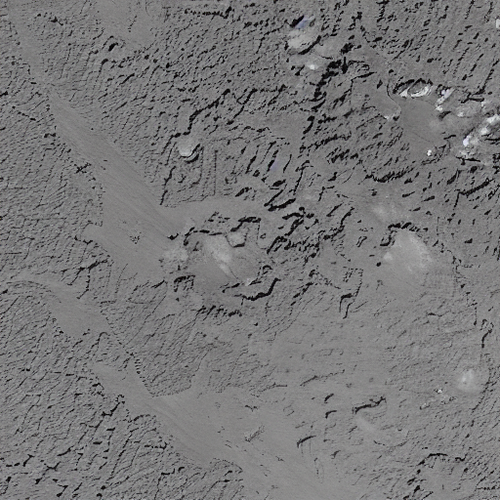}} 
& original prompt &  \small\input{images/qualitative_results/remote_sensing/sdv1.5/12.txt}    \\
& llava caption   & \small\input{images/qualitative_results/remote_sensing/sdv1.5/12_llava.txt}    \\
& llama caption   & \small\input{images/qualitative_results/remote_sensing/sdv1.5/12_llama32.txt}    \\
& molmo caption   & \small\input{images/qualitative_results/remote_sensing/sdv1.5/12_molmo.txt}    \\
\bottomrule
    \end{tabular}}
\end{figure*}

\begin{figure*}[tbp!]
    \centering
    \caption{Example of a generated Texture image with PixArt-$\Sigma$ and the caption generated}
    \label{fig-texture-pixart}
    \resizebox{0.8\textwidth}{!}{
    \begin{tabular}{p{0.2\textwidth}|p{0.15\textwidth}|p{0.65\textwidth}}
\toprule

\multirow{4}{*}{\includegraphics[width=0.2\textwidth]{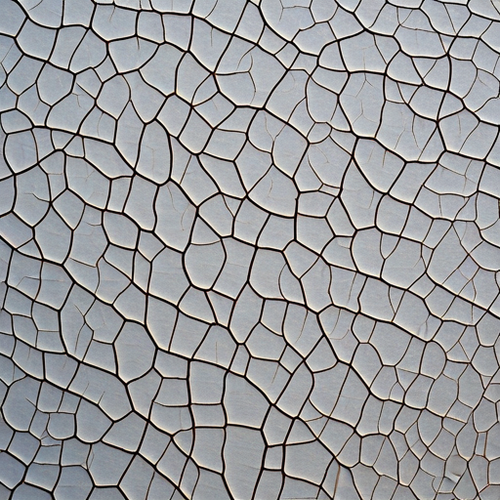}} 
& original prompt &  \small\input{images/qualitative_results/texture/pixart/7.txt}    \\
& llava caption   & \small\input{images/qualitative_results/texture/pixart/7_llava.txt}    \\
& llama caption   & \small\input{images/qualitative_results/texture/pixart/7_llama32.txt}    \\
& molmo caption   & \small\input{images/qualitative_results/texture/pixart/7_molmo.txt}    \\
\bottomrule
    \end{tabular}}
\end{figure*}

\begin{figure*}[tbph!]
    \centering
    
    \caption{Example of a generated Texture image with SDXL and the caption generated }
    \label{fig-texture-sdxl}
    \resizebox{0.8\textwidth}{!}{
    \begin{tabular}{p{0.2\textwidth}|p{0.15\textwidth}|p{0.65\textwidth}}
\toprule
\multirow{4}{*}{\includegraphics[width=0.2\textwidth]{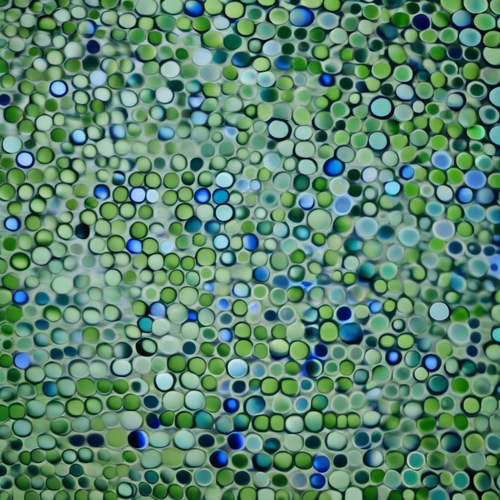}} 
& original prompt &  \small\input{images/qualitative_results/texture/sdxl/7.txt}    \\
& llava caption   & \small\input{images/qualitative_results/texture/sdxl/7_llava.txt}    \\
& llama caption   & \small\input{images/qualitative_results/texture/sdxl/7_llama32.txt}    \\
& molmo caption   & \small\input{images/qualitative_results/texture/sdxl/7_molmo.txt}    \\
\bottomrule
    \end{tabular}}
\end{figure*}

\begin{figure*}[tbp!]
    \centering
    \caption{Example of a generated Texture image with SDXS and the caption generated }
    \label{fig-texture-sdxs}
    \resizebox{0.8\textwidth}{!}{
    \begin{tabular}{p{0.2\textwidth}|p{0.15\textwidth}|p{0.65\textwidth}}
\toprule
\multirow{4}{*}{\includegraphics[width=0.2\textwidth]{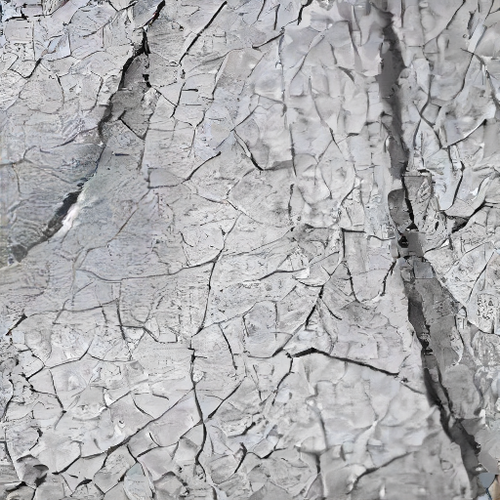}} 
& original prompt &  \small\input{images/qualitative_results/texture/sdxs/7.txt}    \\
& llava caption   & \small\input{images/qualitative_results/texture/sdxs/7_llava.txt}    \\
& llama caption   & \small\input{images/qualitative_results/texture/sdxs/7_llama32.txt}    \\
& molmo caption   & \small\input{images/qualitative_results/texture/sdxs/7_molmo.txt}    \\
\bottomrule
    \end{tabular}}
\end{figure*}

\begin{figure*}[tbp!]
    \centering
    
    \caption{Example of a generated Texture image with SD1.5 and the caption generated }
    \label{fig-texture-sdv1.5}
    \resizebox{0.8\textwidth}{!}{
    \begin{tabular}{p{0.2\textwidth}|p{0.15\textwidth}|p{0.65\textwidth}}
\toprule
\multirow{4}{*}{\includegraphics[width=0.2\textwidth]{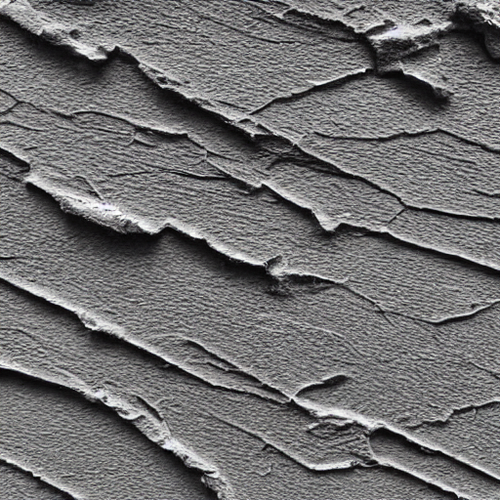}} 
& original prompt &  \small\input{images/qualitative_results/texture/sdv1.5/7.txt}    \\
& llava caption   & \small\input{images/qualitative_results/texture/sdv1.5/7_llava.txt}    \\
& llama caption   & \small\input{images/qualitative_results/texture/sdv1.5/7_llama32.txt}    \\
& molmo caption   & \small\input{images/qualitative_results/texture/sdv1.5/7_molmo.txt}    \\
\bottomrule
    \end{tabular}}
\end{figure*}

\begin{figure*}[ht]
    \centering
    \begin{subfigure}{0.45\textwidth}
        \centering
        \includegraphics[width=\textwidth]{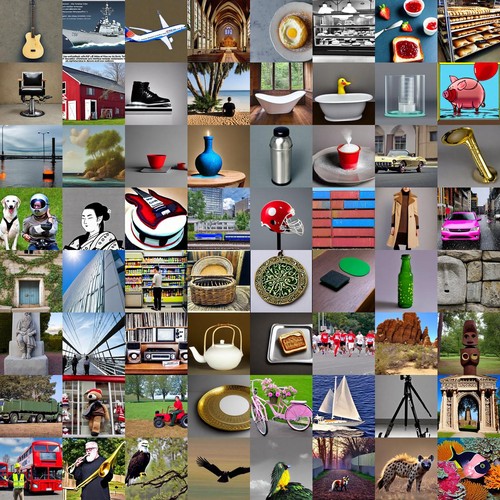}
        \caption*{Stable Diffusion 1.5}
    \end{subfigure}
    \begin{subfigure}{0.45\textwidth}
        \centering
        \includegraphics[width=.99\linewidth]{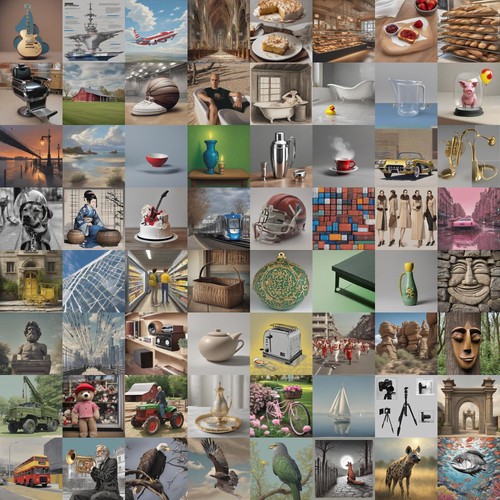}
        \caption*{SDXL}
    \end{subfigure}\\
    \begin{subfigure}[t]{0.45\textwidth}
        \centering
        \includegraphics[width=.99\textwidth]{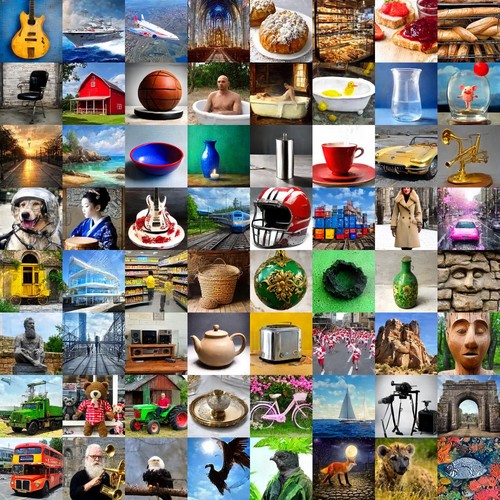}
        \caption*{SDXS}
    \end{subfigure}
    \begin{subfigure}[t]{0.45\textwidth}
        \centering
        \includegraphics[width=.99\linewidth]{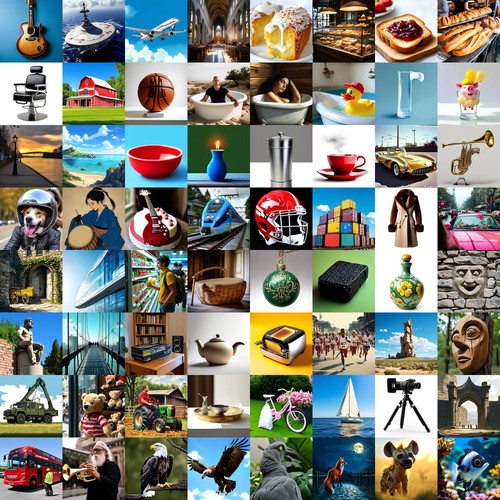}
        \caption*{PixArt $\Sigma$}
    \end{subfigure}
\caption{\textbf{Normal} images with different models. The same prompts are used to generate the images with the four models}
    \label{fig:example_generated_images}
\end{figure*}

\begin{figure*}[ht]
    \centering
    \begin{subfigure}{0.45\textwidth}
        \centering
        \includegraphics[width=\textwidth]{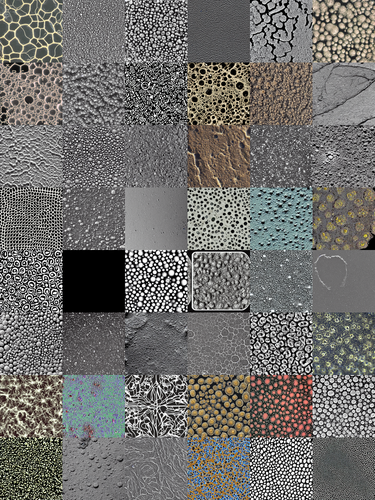}
        \caption*{Stable Diffusion 1.5}
    \end{subfigure}
    \begin{subfigure}{0.45\textwidth}
        \centering
        \includegraphics[width=.99\linewidth]{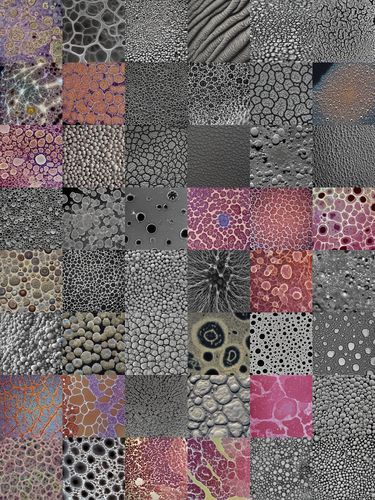}
        \caption*{SDXL}
    \end{subfigure}\\
    \begin{subfigure}[t]{0.45\textwidth}
        \centering
        \includegraphics[width=.99\textwidth]{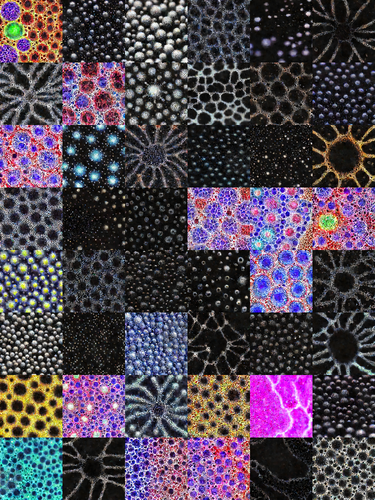}
        \caption*{SDXS}
    \end{subfigure}
    \begin{subfigure}[t]{0.45\textwidth}
        \centering
        \includegraphics[width=.99\linewidth]{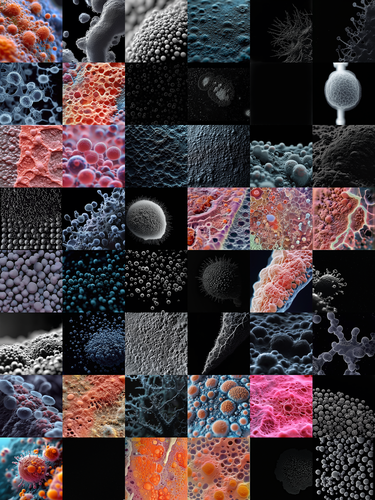}
        \caption*{PixArt $\Sigma$}
    \end{subfigure}
    \caption{\textbf{Microscopic} images with different models. The same prompts are used to generate the images with the four models}
    \label{fig:example_generated_images_micro}
\end{figure*}

\begin{figure*}[ht]
    \centering
    \begin{subfigure}{0.45\textwidth}
        \centering
        \includegraphics[width=\textwidth]{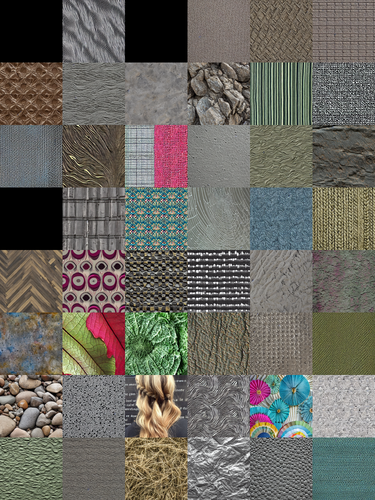}
        \caption*{Stable Diffusion 1.5}
    \end{subfigure}
    \begin{subfigure}{0.45\textwidth}
        \centering
        \includegraphics[width=.99\linewidth]{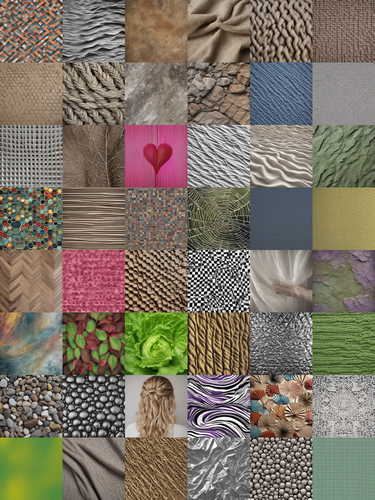}
        \caption*{SDXL}
    \end{subfigure}\\
    \begin{subfigure}[t]{0.45\textwidth}
        \centering
        \includegraphics[width=.99\textwidth]{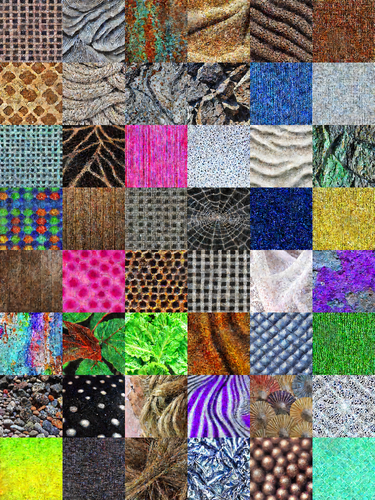}
        \caption*{SDXS}
    \end{subfigure}
    \begin{subfigure}[t]{0.45\textwidth}
        \centering
        \includegraphics[width=.99\linewidth]{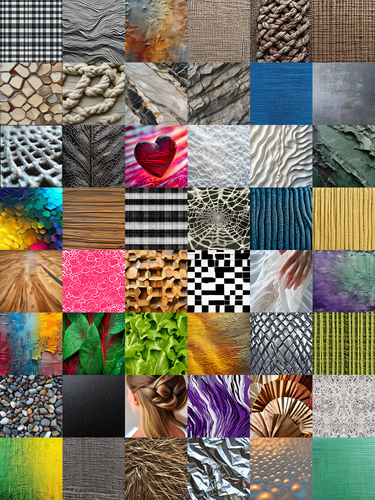}
        \caption*{PixArt $\Sigma$}
    \end{subfigure}
    \caption{\textbf{Texture} images with different models. The same prompts are used to generate the images with the four models}
    \label{fig:example_generated_images_texture}
\end{figure*}

\begin{figure*}[ht]
    \centering
    \begin{subfigure}{0.45\textwidth}
        \centering
        \includegraphics[width=\textwidth]{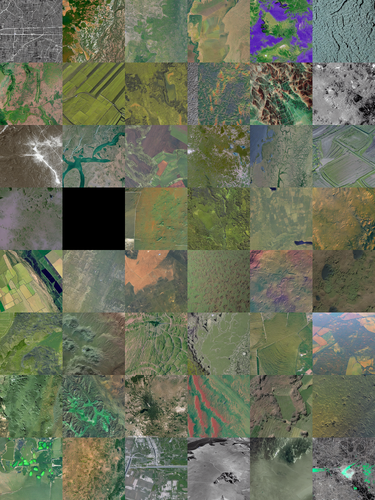}
        \caption*{Stable Diffusion 1.5}
    \end{subfigure}
    \begin{subfigure}{0.45\textwidth}
        \centering
        \includegraphics[width=.99\linewidth]{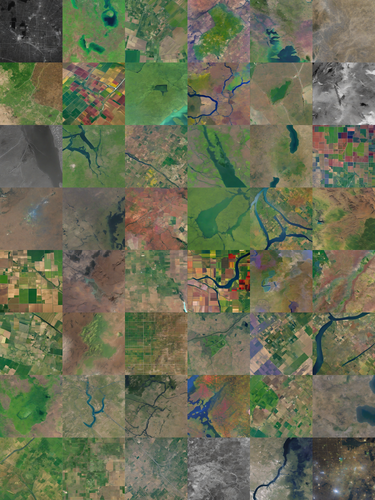}
        \caption*{SDXL}
    \end{subfigure}\\
    \begin{subfigure}[t]{0.45\textwidth}
        \centering
        \includegraphics[width=.99\textwidth]{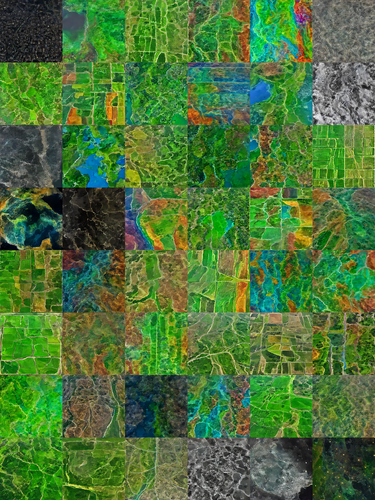}
        \caption*{SDXS}
    \end{subfigure}
    \begin{subfigure}[t]{0.45\textwidth}
        \centering
        \includegraphics[width=.99\linewidth]{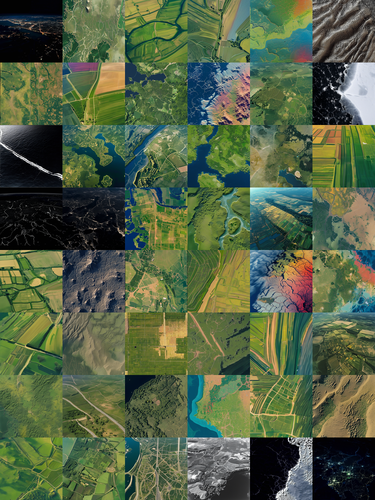}
        \caption*{PixArt $\Sigma$}
    \end{subfigure}
    \caption{\textbf{Remote Sensing} images with different models. The same prompts are used to generate the images with the four models}
    \label{fig:example_generated_images_remote}
\end{figure*}

In this section, we present qualitative results from the images generated and the captions generated from our studies.
In \cref{fig-microscopic-pixart}
\cref{fig-microscopic-sdxl}
\cref{fig-microscopic-sdxs}
\cref{fig-microscopic-sdv1.5}
, we present generated captions of microscopic images. 
In \cref{fig-remote_sensing-pixart}
\cref{fig-remote_sensing-sdxl}
\cref{fig-remote_sensing-sdxs}
\cref{fig-remote_sensing-sdv1.5}
, we present generated captions of Remote Sensing images.
In \cref{fig-texture-pixart}
\cref{fig-texture-sdxl}
\cref{fig-texture-sdxs}
\cref{fig-texture-sdv1.5}
, we present generated captions of Texture images. The \cref{fig:example_generated_images} presents examples of images generated using the same prompt in different T2I models. 
\cref{fig:example_generated_images_micro} presents examples of \textbf{Microscopic} images generated using the same prompt in different T2I models.
\cref{fig:example_generated_images_texture} presents examples of \textbf{Texture} images generated using the same prompt in different T2I models.
\cref{fig:example_generated_images_remote} presents examples of \textbf{Remote Sensing} images generated using the same prompt in different T2I models.


\end{document}